\documentclass[preprint,12pt]{elsarticle}

\usepackage{amssymb}
\usepackage{color}
\usepackage{hyperref}
\usepackage{amsthm}
\usepackage{amsmath}
\usepackage{booktabs}
\usepackage{xcolor}
\usepackage{multirow}
\usepackage{multicol}
\usepackage{makecell}
\usepackage{graphicx}
\usepackage{subcaption} 
\usepackage{algpseudocode}
\usepackage{algorithm}
\usepackage{float}

\journal{GIScience \& Remote Sensing}

\begin{document}

\begin{frontmatter}


\title{An Efficient Remote Sensing Super Resolution Method Exploring Diffusion Priors and Multi-Modal Constraints for Crop Type Mapping}

\author[inst1]{Songxi Yang}

\affiliation[inst1]{organization={Department of Geography},%Department and Organization
            addressline={University of Wisconsin-Madison}, 
            city={Madison},
            postcode={53705}, 
            state={WI},
            country={USA}}

\author[inst1]{Tang Sui}

\author[inst1]{Qunying Huang\corref{cor1}}
\cortext[cor1]{Correspondence to: Qunying Huang (E-mail: qhuang46@wisc.edu)}

\begin{abstract}
Super resolution offers a way to harness medium- even low-resolution but historically valuable remote sensing image archives. Generative models, especially diffusion models, have recently been applied to remote sensing super resolution (RSSR), yet several challenges exist. First, diffusion models are effective but require expensive training-from-scratch resources and have slow inference speeds. Second, current methods have limited utilization of auxiliary information as real-world constraints to reconstruct scientifically realistic images. Finally, most current methods lack evaluation on downstream tasks. In this study, we present a efficient LSSR framework for RSSR, supported by a new multi-modal dataset of paired 30 m Landsat-8 and 10 m Sentinel-2 imagery. Built on frozen pretrained Stable Diffusion, LSSR integrates cross-modal attention with auxiliary knowledge (Digital Elevation Model, land cover, month) and Synthetic Aperture Radar guidance, enhanced by adapters and a tailored Fourier–Normalized Difference Vegetation Index (NDVI) loss to balance spatial details and spectral fidelity. Extensive experiments demonstrate that LSSR significantly improves crop boundary delineation and recovery, achieving state-of-the-art performance with Peak Signal-to-Noise Ratio/Structural Similarity Index Measure of 32.63/0.84 (RGB) and 23.99/0.78 (IR), and the lowest NDVI Mean Squared Error (0.042), while maintaining efficient inference (0.39 sec/image). Moreover, LSSR transfers effectively to NASA Harmonized Landsat and Sentinel-2 (HLS) super-resolution, yielding more reliable crop classification (F1: 0.86) than Sentinel-2 (F1: 0.85). These results highlight the potential of RSSR to advance precision agriculture.
\end{abstract}

% \includegraphics[width=\linewidth]{figures/gra_abs.png}

\begin{highlights}
\item A multi-modal real-world remote sensing super resolution dataset is built with paired 30 m Landsat-8 and 10 m Sentinel-2 imagery, supplemented with DEM, land cover, temporal metadata, and SAR observations.
\item A novel and efficient diffusion-based remote sensing super resolution framework LSSR is proposed, leveraging Stable Diffusion priors, cross-modal attention with physical-world constraints (DEM, land cover, month), and SAR-guided fusion.
\item LSSR achieves competitive generative results while costing only slight trainable parameter and inference time increments.
\item LSSR demonstrates strong transferability to NASA HLS data, enabling reliable crop type mapping.

%% \item The corresponding LSSR dataset is publicly available for both super resolution and application in precision agriculture.

\end{highlights}


\end{frontmatter}


\section{Introduction}
\label{sec:intro}
Remote sensing (RS) provides various and long-term observations of the Earth's surface \cite{campbell2011introduction}. However, the spatial and spectral resolutions of historical satellite archives (e.g., Landsat \cite{markham2004landsat} and MODIS \cite{wang2012impact}) are generally insufficient to meet the requirements of recent fine-grained applications, due to limitations in optical systems, sensor degradation, and the high cost associated with acquiring high-resolution imagery \cite{wang2022comprehensive}. However, these series are important for long-term applications such as environmental monitoring under climate change, land use/cover change over decades, and climate modeling using historical archives, that have been used by a variety of Environmental Sciences disciplines \cite{wang2022comprehensive}. 

Among various RS datasets, the Harmonized Landsat and Sentinel-2 (HLS) product provides both temporally dense (every 2-3 days), moderate spatial resolution (30m) and radiometrically consistent observations by fusing surface reflectance data from the Landsat-8 Operational Land Imager (OLI) and Sentinel-2 MultiSpectral Instrument (MSI) \cite{claverie2018hls}. HLS includes two products, S30 and L30, derived from Sentinel-2 and Landsat input, respectively, which enable high-frequency monitoring of Earth's surface, supporting applications such as crop yield prediction, land cover classification, and phenological analysis. However, although Sentinel-2 provides some bands at 10m, a key limitation of the HLS product retains the moderate spatial resolution (30m). This spatial limitation hampers fine-grained monitoring and reduces the effectiveness of downstream applications that require detailed spatial structures, such as field-level crop mapping or small-scale land use change detection. Therefore, it is of paramount significance to develop algorithms to improve the spatial and spectral quality of these satellite images.

Remote Sensing Super Resolution (RSSR) aims to reconstruct a high-resolution (HR) image by enhancing the spatial and/or spectral quality of the low-resolution (LR) image counterpart \cite{wang2020deep}, thereby providing an opportunity to improve the spatial resolution of HLS data. A wide range of SR methods have been developed to tackle in the RS field, ranging from classical interpolation techniques to advanced Deep Learning (DL) based approaches \cite{al2024single}. The earliest Convolutional Neural Network (CNN-based) SR model, SRCNN \cite{dong2015srcnn}, is one of the earliest DL-based SR models, introducing a simple three-layer convolutional architecture that significantly improved performance over traditional interpolation-based and reconstruction-based methods. VDSR \cite{kim2016vdsr} builds on this by employing deeper, 10-layer residual learning \cite{he2016deep}, allowing for faster convergence and improved accuracy. EDSR \cite{lim2017edsr} further optimizes the common residual structure by removing batch normalization, enabling even deeper networks and achieving better results on many benchmarks. These CNN-based models have laid a strong foundation for transferring SR techniques into RS imagery. For instance, GEOSR \cite{hecht2008geosr} integrates and adapts these classical models for RS tasks, demonstrating the utility of these architectures in domain-specific SR. 

In addition to CNN-based models, attention mechanisms have also been introduced for capturing global and local context \cite{vaswani2017attention}. Examples include the Multi-scale Attention Network (MAN) \cite{wang2024man},  which integrates attention modules across multiple scales to enhance the representation of the network, thus achieving superior performance on many SR benchmarks. More recently, Transformer-based architectures such as SwinIR \cite{liang2021swinir} have shown promising results by employing hierarchical self-attention to balance efficiency and representational power. 

More recently, generative models have been reshaping the computer vision domain. Over the past several years, we’ve seen the rise of advanced models like Generative Adversarial Network (GAN) and diffusion architectures \cite{wang2020deep}. A typical Generative Adversarial Network (GAN) consists of two models: a discriminator and a generator \cite{goodfellow2014gan}. A discriminator estimates the probability of a given sample coming from the real dataset. It works as a critic and is optimized to distinguish the fake samples from the real ones. A generator outputs synthetic samples given a noise variable input. It is trained to capture the real data distribution so that its generative samples can be as real as possible. This competitive game between two models motivates both to improve their functionalities. SRGAN \cite{ledig2017srgan} was the first to introduce adversarial loss for SR, pushing the output towards more realistic textures and sharper details. Building on this, ESRGAN (Enhanced SRGAN) \cite{wang2018esrgan} further refines the SRGAN architecture with residual-in-residual dense blocks and a perceptual loss. These improvements lead to both better perceptual quality and higher quantitative metrics. More recent works have extended ESRGAN to domain specific tasks—for instance, applying ESRGAN to infrared (IR) image super-resolution \cite{vassilo2021infrared}, demonstrating its adaptability to diverse data modalities beyond the natural color (RGB) domain. However, GAN-based approaches often suffer from artifacts, convergence instability, and mode collapse, where the model generates overly similar textures instead of capturing diverse high-resolution details \cite{zhang2018convergence}.

Following the limitations of GAN-based models, diffusion-based generative models have emerged as an alternative to adversarial learning. Unlike GANs that rely on adversarial objectives, diffusion models define a Markov chain of diffusion steps to slowly add random noise to degrade original data and then learn to reverse the diffusion process to construct desired data samples from the noise \cite{ho2020ddpm}. Methods have been proposed to make the process much faster, such as denoising diffusion implicit models (DDIMs), but the sampling process is still slower than GANs \cite{song2020ddim}. To address this, new methods aim to accelerate inference. Recently, studies demonstrate that the diffusion priors, embedded in pretrained Stable Diffusion \cite{rombach2022stablediffusion}, can be applied to various downstream content creation tasks, offering adaptability and competitive performance \cite{he2025diffusionsurvey}. For example, StableSR \cite{wang2024stablesr} adds trainable spatial feature transform layers to exploit Stable Diffusion priors. Moreover, Pixel-level and Semantic-level Adjustable Super-resolution (PiSA-SR) \cite{sun2025pisasr} is a dual approach, characterizing pixel-level and semantic-level information, achieving results in both quality and efficiency. 

In RSSR area, DiffusionSat \cite{khanna2023diffusionsat} is a notable example. It trains Stable Diffusion from scratch and leverages RS image metadata (longitude, latitude, ground-sampling distance, cloud cover, timestamp) as additional embeddings, enabling effective RSSR. Moreover, An adaptive
semantic-enhanced DDPM (ASDDPM) \cite{sui2024adaptive} introduces an Adaptive Detail Fusion Transformer Decoder (ADTD) to enhance semantic representation and a residual feature fusion strategy. Experiments are conducted on four datasets, including one Landsat–Sentinel paired dataset OLI2MSI \cite{wang2021multisensor}. The Efficient Variance Attention-enhanced Diffusion Model (EVADM) \cite{lu2024effective} introduces a Variance-Average-Spatial Attention (VASA) mechanism to improve detail recovery in crop field aerial image SR. The authors built a large-scale CropSR dataset and two real-matched testing datasets CropSR-Ortho and CropSR-Fixed-Point from aerial photography. Furthermore, the downstream case study shows that EVADM achieved more reliable recognition of rice growth stages compared with EDSR and RealESRGAN \cite{lu2024effective}.

Despite these advancements, existing diffusion-based RSSR methods often suffer from three limitations: challenges in balancing reconstruction effectiveness with inference efficiency in real-world RS data, limited utilization of auxiliary information as real-world constraints, and the scarcity of downstream task evaluations after RSSR. To address these gaps, we propose a novel diffusion-based framework (LSSR) that explicitly incorporates multispectral RS characteristics and domain-specific priors. We summarized several innovations of our study: 
\begin{itemize}
    \item Built a multi-modal RSSR dataset comprising paired 30 m Landsat-8 and 10 m Sentinel-2 images, supplemented with auxiliary information such as Digital Elevation Model (DEM), land cover types, temporal metadata, and synthetic-aperture radar (SAR).
    \item Extended the PiSA-SR method by developing a Cross-attention Knowledge Constraint Module that injects geophysical and temporal features into the Stable Diffusion latent space.
    \item Designed a SAR-guided Fusion Block that integrates structural features to refine textures.
    \item Proposed a spectral–frequency joint loss function, which combines a Fourier–Vegetation Index hybrid loss to enhance spectral fidelity.
    \item Transferred and evaluated the LSSR model performance by downstream crop type mapping on NASA HLS images.
\end{itemize}

\section{Data}
\label{sec:data}

\subsection{LSSR Data Collection and Preprocessing}

We collected a total of 1,853 paired samples of 30 m Landsat-8 \cite{usgslandsat8toa} and 10 m Sentinel-2 \cite{sentinel2toa} images, along with corresponding 10 m DEM from the USGS 3D Elevation Program \cite{usgs3dep}, 10 m land cover types from Dynamic World \cite{brown2022dynamic}, and 10 m Sentinel-1 SAR, including Vertical–Vertical (VV), and Vertical–Horizontal (VH) polarization \cite{sentinel1} (referred to as LSSR dataset). Data preprocessing \cite{berra2024harmonized} was conducted on Google Earth Engine (GEE) platform \cite{gorelick2017gee}, including general filtering, cloud and shadow screening, inter-sensor band adjustment, and atmospheric correction \cite{berra2024harmonized}. The Landsat-8, Sentinel-2, and Sentinel-1 SAR GRD images were not on the exact same date due to different revisit cycles, and we paired them based on the closest acquisition dates within a 7-day window to maximum the temporal consistency. To ensure spectral consistency, 2 shortwave infrared bands of Sentinel-2 were resampled to 10 m in GEE. This allows the model to process all inputs on a uniform spatial grid. All datasets were geo-registered to the same projection as Sentinel-2 before patch extraction. Each set of samples covers a spatial extent of 64 × 64 pixels at 30 m for Landsat-8, 192 × 192 pixels at 10 m for Sentinel-2 and Sentinel-1 SAR, ensuring spatial alignment. For model training, the dataset was split into 1,377 training pairs and 476 testing pairs. The selected spectral bands are listed in Table \ref{tab:l8s2hls}.

\begin{table}[ht]
\centering
\caption{Selected Spectral Band Number of Landsat-8, Sentinel-2, and HLS}
\label{tab:l8s2hls}
\resizebox{\textwidth}{!}{
\begin{tabular}{cccccc}
\hline
\textbf{Band Name} & \textbf{Landsat-8 Band} & \textbf{Sentinel-2 Band} & \textbf{HLSL30 Band} & \textbf{HLSS30 Band} \\
\hline
Blue & B2 & B2 & B02 & B02 \\
Green & B3 & B3 & B03 & B03 \\
Red & B4 & B4 & B04 & B04 \\
NIR & B5 & B8 & B05 & B08 \\
SWIR 1 & B6 & B11 & B06 & B11 \\
SWIR 2 & B7 & B12 & B07 & B12 \\
\hline
\end{tabular}
}
\end{table}

\begin{figure}[!htbp]
  \centering
  \includegraphics[
    width=0.9\linewidth,
    trim=0mm 10mm 0mm 10mm, clip
  ]{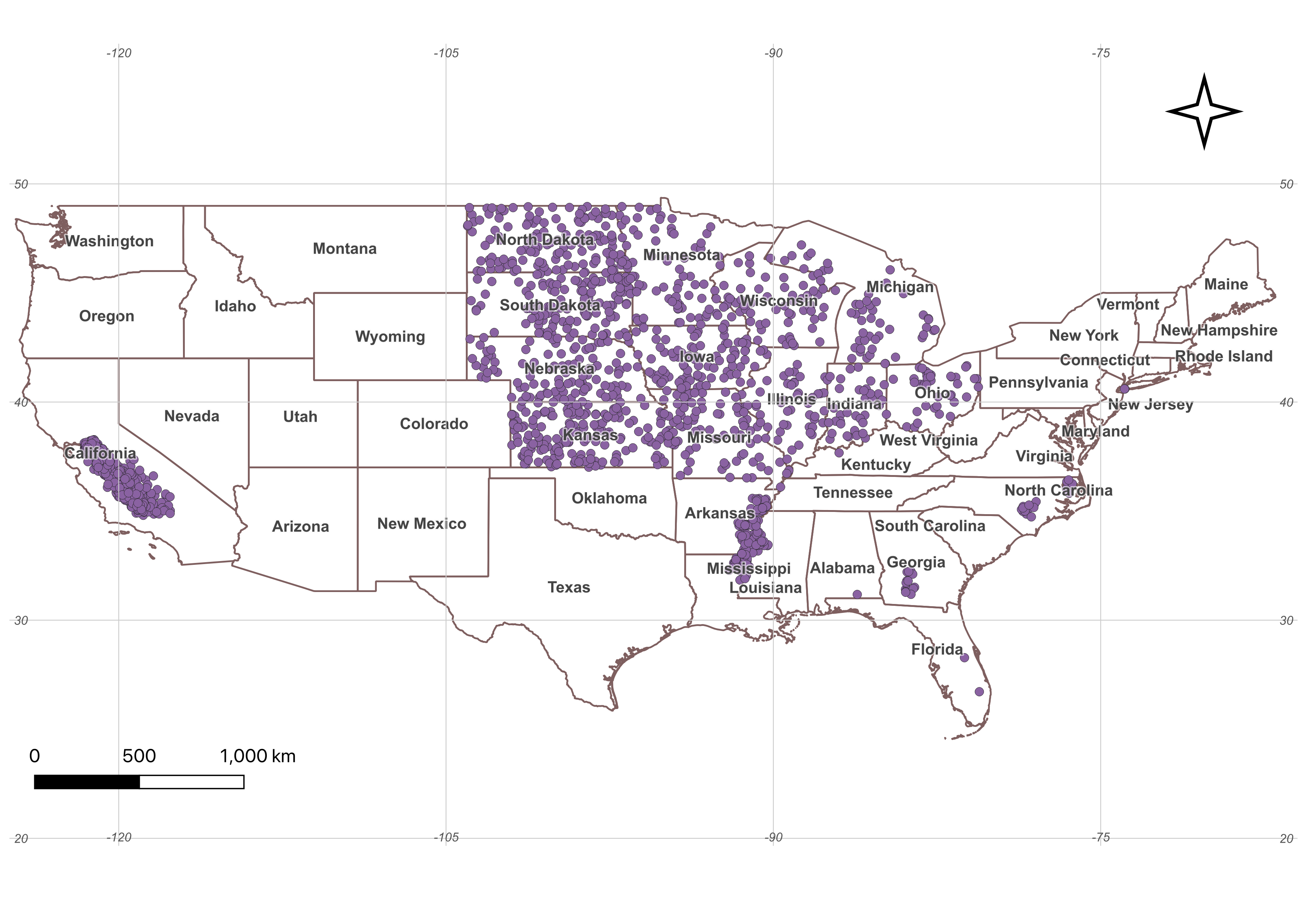}
  \caption{Geospatial Distribution of the LSSR Dataset.}
  \label{fig:geo}
\end{figure}

\begin{figure}[htbp]
    \centering

    \begin{subfigure}{\textwidth}
        \centering
        \includegraphics[width=\linewidth]{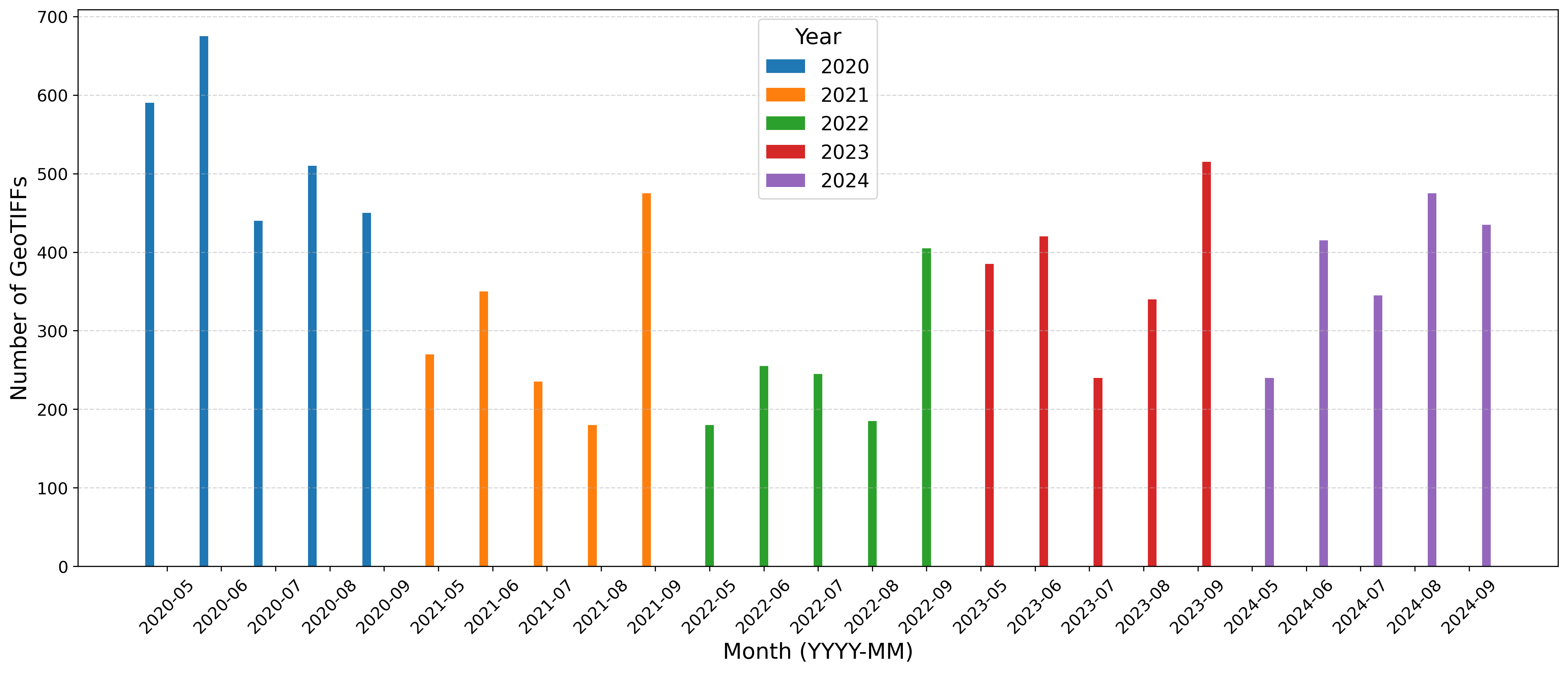}
        \caption{Monthly temporal coverage of the LSSR dataset.}
        \label{fig:temporal}
    \end{subfigure}

    \vspace{1.5em}

    \begin{subfigure}{0.8\textwidth}
        \centering
        \includegraphics[width=\linewidth]{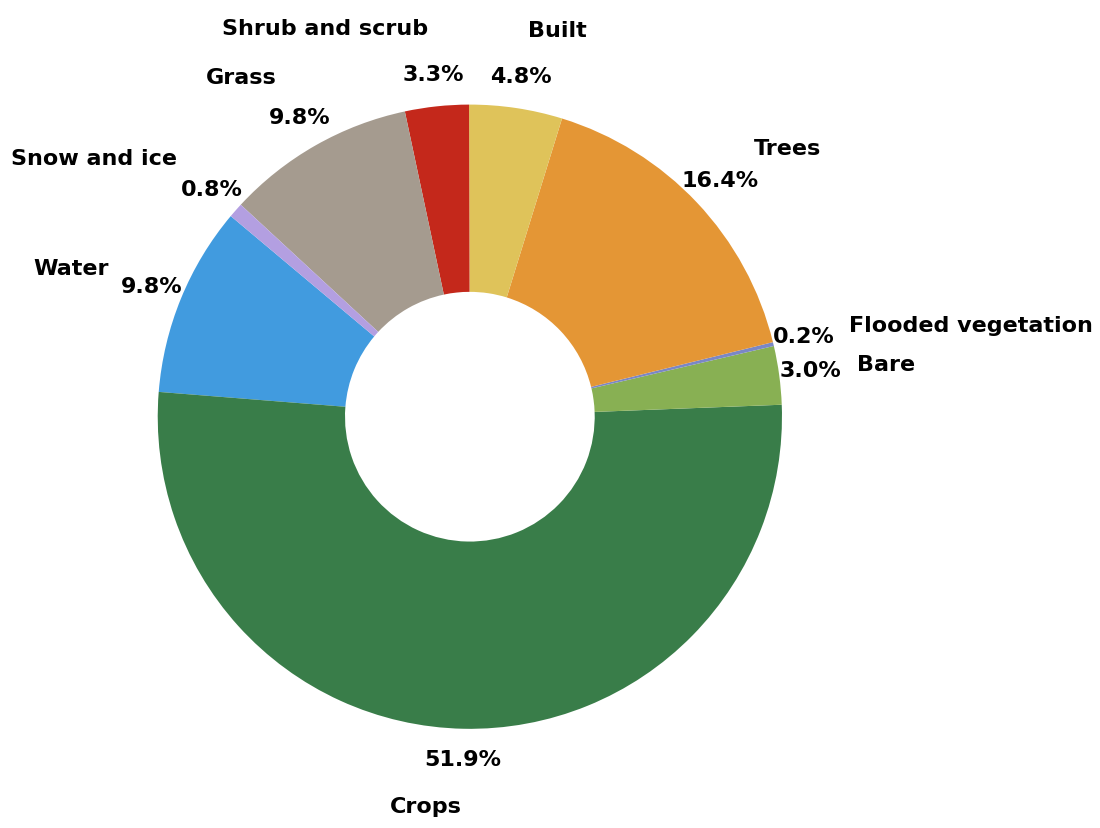}
        \caption{Land cover distribution of the LSSR dataset.}
        \label{fig:landcover}
    \end{subfigure}

    \caption{Statistical overview of the LSSR dataset.}
    \label{fig:lssr_stats}
\end{figure}

Table \ref{tab:datainfo} shows detailed regions where the dataset is coming from geographically. We further illustrate the spatial, temporal, and land-cover coverages of the dataset in Figure \ref{fig:geo} and \ref{fig:lssr_stats}. The dataset comprises paired images collected across diverse agricultural regions in the United States, with a primary focus on the California Central Valley, Midwest, and Southeast. In Figure \ref{fig:geo}, each data point represents a unique georeferenced tile. Temporally, the dataset spans five years (2020–2024), with image acquisitions concentrated in the growing season from May to October (Figure \ref{fig:temporal}). Monthly counts indicate consistent seasonal coverage. This temporal consistency helps the model to monitor vegetation phenology and crop development over multiple years.

Based on Figure \ref{fig:landcover}, cropland constitutes over half (51.88\%) of the entire dataset, substantially outweighing other land cover classes. This design choice reflects the dataset’s targeted application: enhancing spatial resolution in agricultural regions for downstream tasks such as crop type mapping, phenology monitoring, and crop yield estimation. Moreover, the inclusion of ancillary classes (e.g., trees, grass, water, bare soil) ensures that models trained on LSSR maintain robustness across mixed land cover scenes while preserving a strong focus on cropland structures. This balance enables the development of SR algorithms that are both domain-adaptive and crop-sensitive, aligning with the practical requirements of real-world agricultural applications.

\begin{figure*}[htbp]
  \centering
  \includegraphics[width=\linewidth]{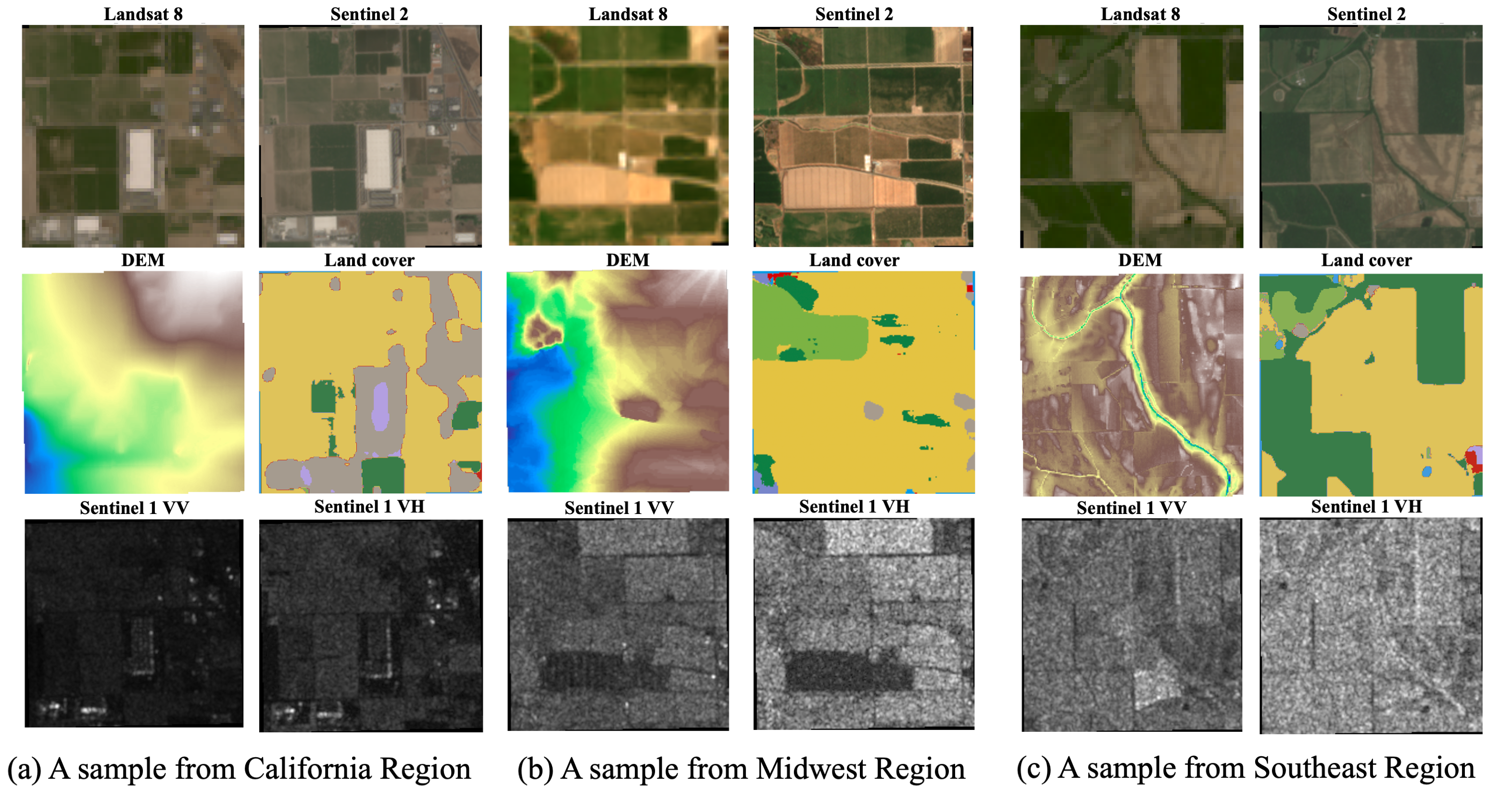}
  \caption{Sample Pairs from the LSSR Dataset.}
  \label{fig:sample}
\end{figure*}

\begin{table}[ht]
\centering
\caption{LSSR Data Information}
\label{tab:datainfo}
\resizebox{\textwidth}{!}{
\begin{tabular}{lccc}
\hline
\textbf{Region} & \textbf{Time (YYYY/MM)} & \textbf{Data Pairs} \\
\hline
Lower Midwest, United States & 2020/05 - 2024/09 &  539 \\
Upper Midwest, United States & 2020/05 - 2024/09 &  535 \\
California Central Valley, United States & 2020/05 - 2024/09 & 581 \\
Southeastern, United States & 2020/05 - 2024/09 & 198 \\
\hline
\end{tabular}
}
\end{table}

Figure \ref{fig:sample} shows representative samples from California Central Valley, Midwest, and Southeast, respectively. Each sample consists of co-registered Landsat-8 and Sentinel-2 optical images, along with derived auxiliary layers including a DEM, a land cover classification map, and Sentinel-1 images. We collected samples from heterogeneous agricultural landscapes across regions, ranging from the highly structured irrigation grids in California to the mixed vegetation and topographic variation in the Southeast.

\subsection{HLS Data Collection and Preprocessing}

We collected 129 training samples from Dane County and 75 testing samples from Columbia County, Wisconsin, USA, to evaluate the performance of LSSR in a downstream crop classification task. Each sample consists of 30 m HLS imagery from June, July, and August, along with auxiliary data including a 10 m DEM, a 10 m land cover map, and 10 m Sentinel-1 images for LSSR guidance. For comparison, we also include 30 m Landsat-8 and 10 m Sentinel-2 images for direct crop type classification benchmarking. Spectral bands across sensors are matched according to Table \ref{tab:l8s2hls}. The crop type labels are obtained from the 30 m USDA NASS Cropland Data Layer (CDL), which provides annual nationwide crop classification. All labels are reprojected to same coordinate system to ensure label consistency across Landsat-8, Sentinel-2, and HLS inputs. The 30 m HLS imagery is super-resolved by the proposed LSSR method to 10 m resolution, and classification performance is compared using four inputs: 10 m super-resolved HLS, original 30 m HLS, 30 m Landsat-8, and 10 m Sentinel-2.

\section{Method}
\label{sec:methods}

\subsection{LSSR Method}

\subsubsection{Diffusion Models and Parameter-Efficient Fine-Tuning}

Diffusion models, inspired by non-equilibrium thermodynamics, consist of two stages: a forward (noising) process and a reverse (denoising) process. In the forward process, noise is progressively added to the data, while in the reverse process, a learned noise prediction model estimates and removes the noise step by step to recover the original features \cite{ho2020ddpm}.

In the forward diffusion process, given a clean image $x_0$, the forward Markov process gradually adds Gaussian noise:
\begin{equation}
q(x_t \mid x_{t-1}) = \mathcal{N}\!\big(\sqrt{1-\beta_t}\,x_{t-1},\,\beta_t \mathbf{I}\big), \quad t=1,\dots,T ,
\end{equation}
where $\{\beta_t\}$ is the noise schedule. This yields a closed-form expression:
\begin{equation}
q(x_t \mid x_0) = \mathcal{N}\!\big(\sqrt{\bar\alpha_t}\,x_0,\,(1-\bar\alpha_t)\mathbf{I}\big), \quad 
\alpha_t = 1-\beta_t,\;\bar\alpha_t=\prod_{i=1}^t \alpha_i .
\end{equation}

During the reverse process, the goal is to learn a parameterized model to approximate the reverse transition:
\begin{equation}
p_\theta(x_{t-1}\mid x_t,\mathbf{c})=\mathcal{N}\!\big(\mu_\theta(x_t,t,\mathbf{c}),\,\sigma_t^2\mathbf{I}\big),
\end{equation}
where $\mathbf{c}$ denotes the conditioning information (e.g., a LR image).  
A common parameterization is noise prediction:
\begin{equation}
\mu_\theta(x_t,t,\mathbf{c})=\frac{1}{\sqrt{\alpha_t}}\left(x_t-\frac{\beta_t}{\sqrt{1-\bar\alpha_t}}\,\epsilon_\theta(x_t,t,\mathbf{c})\right).
\end{equation}

In the SR field, the model is conditioned on a low-resolution image $y=H(x_0)$, where $H$ is the degradation operator. However, training diffusion models from scratch can be computationally expensive \cite{song2020ddim}. To address this, parameter-efficient fine-tuning (PEFT) methods such as Low-Rank Adaptation (LoRA) have been introduced \cite{hu2022lora}. LoRA reduces the number of trainable parameters by decomposing weight updates into low-rank matrices, enabling efficient adaptation of large pretrained diffusion models to domain-specific tasks like RSSR \cite{sun2025pisasr}.

\subsubsection{LSSR Model Architecture}

\begin{figure*}[h]
  \centering
  \includegraphics[width=\linewidth]{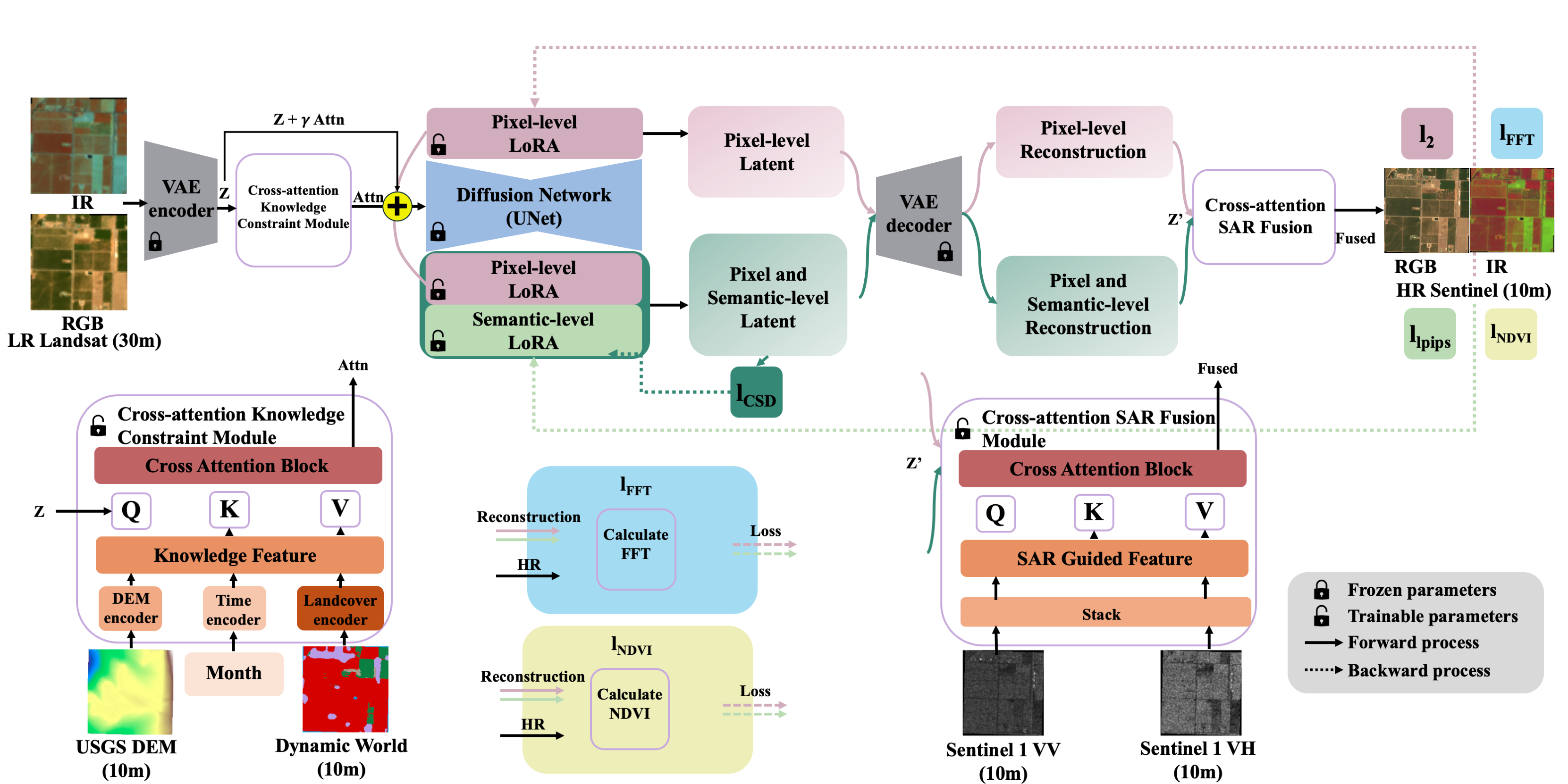}
  \caption{LSSR Model Architecture, modified from \cite{sun2025pisasr}. The Diffusion Network is frozen and fine-tuned through pixel-level and semantic-level LoRA adapters; thus, the LoRA output corresponds to the adapted output features of the Diffusion Network.}
  \label{fig:lssr_architecture}
\end{figure*}

Figure \ref{fig:lssr_architecture} shows the proposed LSSR model architecture. The LSSR consists of: frozen Stable Diffusion, frozen VAE encoder/decoder, trainable dual-branch LoRA modules, a trainable cross-attention knowledge constraint module, a trainable cross-attention SAR fusion module, and loss functions.

First, the LSSR architecture is built upon a pretrained Stable Diffusion model, where both the VAE encoder/decoder and diffusion UNet are kept frozen to leverage strong generative priors \cite{sun2025pisasr}. Following the design of the PiSA-SR model \cite{sun2025pisasr}, given a LR input image, LSSR obtains its latent representation Z through a frozen VAE encoder. This latent is then passed into the diffusion backbone enhanced with two parallel LoRA branches: the pixel-level LoRA branch focuses on fine-grained texture restoration. This output is supervised using a pixel-wise \{$l_2$\} loss against the reference HR image. The pixel and semantic-level LoRA branch introduces both pixel- and semantic-level PEFT. It produces another refined latent for reconstruction, guided by a perceptual loss (LPIPS) and a contrastive semantic distillation loss (CSD) to enhance semantic fidelity and structure alignment. The final reconstructed outputs are decoded by the frozen VAE decoder, sharpening local details and enhancing global consistency.

Furthermore, to incorporate knowledge priors into the latent space, we introduce a cross-attention knowledge constraint mechanism that injects auxiliary information, such as DEM, land cover type, and month index, into the image latent representation. As shown in Figure \ref{fig:lssr_architecture}, each auxiliary modality is first encoded into a consistent feature map through a shallow convolutional encoder (for DEM and land cover) or an embedding layer (for month). These features are then aggregated into a single auxiliary representation:

\begin{equation}
\mathbf{z}_{\text{aux}} = \mathbf{f}_{\text{DEM}} + \mathbf{f}_{\text{LC}} + \mathbf{f}_{\text{Month}}
\end{equation}

To enable interaction in a shared attention space, both the image latent $\mathbf{z}_{\text{img}}$ and auxiliary latent $\mathbf{z}_{\text{aux}}$ are projected into query, key, and value tensors:

\begin{equation}
Q = \text{Proj}(\mathbf{z}_{\text{img}}), \quad K = V = \text{Proj}(\mathbf{z}_{\text{aux}})
\end{equation}

Then, the cross-attention output is computed using multi-head attention:

\begin{equation}
\mathbf{Attn} = \text{Attention}(Q, K, V) = \text{softmax}\left( \frac{Q K^{\top}}{\sqrt{d}} \right) V
\end{equation}

The resulting attended features are projected back and injected into the original latent representation via residual connection:

\begin{equation}
\hat{\mathbf{z}}_{\text{img}} = \mathbf{z}_{\text{img}} + \gamma \cdot \text{Proj}^{-1}(\mathbf{Attn})
\end{equation}

where $\gamma$ is a learnable scalar controlling the strength of auxiliary conditioning automatically during the training process. This mechanism is applied separately to both RGB and IR latent branches to enable modality-aware prior injection.

Finally, to refine reconstructed embedding with structural priors from SAR data, we design a cross-attention SAR fusion module. The module takes a 3-band RGB/IR image and a 2-band VH VV stacked SAR image as inputs and consists of three stages.

First, in the feature projection stage, the RGB/IR image $z'\in \mathbb{R}^{3\times H\times W}$ and the SAR image $I_{sar}\in \mathbb{R}^{2\times H\times W}$ 
are first projected into a shared latent space by shallow convolutional encoders:
\begin{equation}
F_{v} = \phi_{v}(z'), \quad 
F_{sar} = \phi_{sar}(I_{sar}).
\end{equation}

Then, in the cross-attention fusion stage, we employ multi-head cross-attention where RGB/IR features serve as queries and SAR features provide keys and values:
\begin{equation}
\text{Attn}(Q,K,V) = \text{softmax}\!\left(\frac{QK^\top}{\sqrt{d}}\right)V,
\end{equation}
with $Q=W_qF_{v}, \; K=W_kF_{sar}, \; V=W_vF_{sar}$. 
The fused RGB/IR features are obtained as
\begin{equation}
Fused = F_{v} + \gamma \cdot G(F_{sar}) \odot \text{Attn}(Q,K,V),
\end{equation}
where $G(\cdot)$ denotes a gating function that adaptively scales SAR contributions, and $\gamma$ is a learnable global scalar.

Our training objective is designed to jointly optimize low-level pixel fidelity, high-level semantic consistency, and physical world reliability. In addition to a pixel-wise loss, a Learned Perceptual Image Patch Similarity (LPIPS) loss and a CSD loss in the original PiSA-SR architecture \cite{sun2025pisasr}, we also propose a Fast Fourier Transform (FFT) loss function \cite{wang2023multiscalefft, fuoli2021fourierloss}, and a NDVI loss function. Specifically, the total loss function consists of five components:

\begin{equation}
\mathcal{L}_{\text{total}} = \mathcal{L}_{\text{RGB}} + \mathcal{L}_{\text{IR}} 
\end{equation}

\begin{equation}
\mathcal{L}_{\text{RGB}} = \lambda_{\text{2}} \cdot \mathcal{L}_{\text{2}} + \lambda_{\text{lpips}} \cdot \mathcal{L}_{\text{lpips}} + \lambda_{\text{csd}} \cdot \mathcal{L}_{\text{csd}} + \lambda_{\text{fft}} \cdot \mathcal{L}_{\text{fft}} + \lambda_{\text{ndvi}} \cdot \mathcal{L}_{\text{ndvi}}
\end{equation}

\begin{equation}
\mathcal{L}_{\text{IR}} = \lambda_{\text{2}} \cdot \mathcal{L}_{\text{2}} + \lambda_{\text{lpips}} \cdot \mathcal{L}_{\text{lpips}} + \lambda_{\text{csd}} \cdot \mathcal{L}_{\text{csd}} + \lambda_{\text{fft}} \cdot \mathcal{L}_{\text{fft}} + \lambda_{\text{ndvi}} \cdot \mathcal{L}_{\text{ndvi}}
\end{equation}

The pixel-level reconstruction loss is computed as:

\begin{equation}
\mathcal{L}_{\text{2}} = \| \hat{\mathbf{x}}_H^{\text{2}} - \mathbf{x}_H \|_2^2
\end{equation}

where \( \hat{\mathbf{x}}_H^{\text{2}} \) is the reconstructed image from the pixel-level LoRA branch, and \( \mathbf{x}_H \) is the ground-truth high-resolution Sentinel-2 image.

The LPIPS loss measures high-level similarity using a pretrained VGG network:

\begin{equation}
\mathcal{L}_{\text{lpips}} = \text{LPIPS}(\hat{\mathbf{x}}_H^{\text{sem}}, \mathbf{x}_H)
\end{equation}

The CSD loss is applied in latent space, encouraging the semantic-aware branch to better align with the diffusion prediction target. Denoting the predicted and ground truth latents as \( \hat{\mathbf{z}} \) and \( \mathbf{z} \), respectively, we define:

\begin{equation}
\mathcal{L}_{\text{csd}} = \| \mathbf{z} - \text{stopgrad}(\mathbf{z} - \nabla_{\mathbf{z}}) \|_2^2
\end{equation}

where \( \nabla_{\mathbf{z}} \) is the scaled gradient estimated using contrastive noise prediction.

To enhance high-frequency detail reconstruction, we enforce alignment in the frequency domain using the Discrete Fourier Transform (DFT):
\begin{equation}
\mathcal{L}_{\text{fft}} = \| \mathcal{F}(\hat{\mathbf{x}}_H^{\text{sem}}) - \mathcal{F}(\mathbf{x}_H) \|_1
\end{equation}
where \( \mathcal{F}(\cdot) \) denotes the 2D Fourier transform, and \( \| \cdot \|_1 \) is used to emphasize sparsity in spectral differences.

To preserve vegetation semantics, we compute NDVI (Normalized Difference Vegetation Index) from predicted and ground-truth images:
\begin{equation}
\mathcal{L}_{\text{ndvi}} = \| \text{NDVI}(\hat{\mathbf{x}}_H^{\text{sem}}) - \text{NDVI}(\mathbf{x}_H) \|_2^2
\end{equation}
Here, NDVI is computed from red and near-infrared bands as:
\begin{equation}
\text{NDVI} = \frac{B_{\text{NIR}} - B_{\text{R}}}{B_{\text{NIR}} + B_{\text{R}} + \epsilon}
\end{equation}
where \( \epsilon \) is a small constant to avoid division by zero.

The weights \( \lambda_{\text{pix}}, \lambda_{\text{lpips}}, \lambda_{\text{csd}} \), \( \lambda_{\text{fft}} \), \( \lambda_{\text{ndvi}} \) control the contribution of each loss term and are set empirically. The final weights were set to \( \lambda_{\text{pix}}=2.0, \lambda_{\text{lpips}}=1.0, \lambda_{\text{csd}} \)=2.0, \( \lambda_{\text{fft}} \)=1.0, \( \lambda_{\text{ndvi}} \)=20.0. The relative large magnitude of NDVI-based constraint can effectively enforce physical consistency between spectral bands, rather than being overshadowed by pixel- or feature-level losses. The detailed ablation study can be found in Section \ref{sec:ablation}.

\subsubsection{Evaluation Metrics}

We evaluate the reconstruction performance across both RGB and infrared (IR) bands using a comprehensive set of metrics that capture low-level fidelity, perceptual similarity, and semantic consistency. The detailed descriptions are listed in Table~\ref{tab:metrics}.

\begin{table}[H]
\centering
\caption{Summary of Evaluation Metrics in This Study.}
\label{tab:metrics}
\begin{tabular}{p{3cm}p{9cm}}
\toprule
\textbf{Metric} & \textbf{Description} \\
\midrule
\multicolumn{2}{l}{\textit{Super-Resolution Evaluation Metrics}} \\
\midrule
\textbf{PSNR} $\uparrow$ & Peak Signal-to-Noise Ratio for RGB 
 and IR channels, respectively; measures pixel-wise fidelity (in dB). Higher is better. \\

\textbf{SSIM} $\uparrow$ & Structural Similarity Index Measure for RGB 
 and IR channels, respectively; captures texture and structure similarity. Ranges from 0 to 1. \\

\textbf{LPIPS} $\downarrow$ & Learned Perceptual Image Patch Similarity; perceptual metric using deep features. Lower indicates better perceptual similarity. \\

\textbf{FCL} $\downarrow$ & Feature Consistency Loss for RGB 
 and IR channels, respectively; L2 distance between deep feature embeddings (from VGG network). Lower is better. \\

\textbf{SAM} $\downarrow$ & Spectral Angle Mapper; reflects spectral consistency between the reconstructed and reference images. \\

\textbf{NDVI MSE} $\downarrow$ & Mean Squared Error of NDVI between prediction and ground truth; reflects semantic correctness in vegetation information. \\

\textbf{Infer. Time} $\downarrow$ & Average time to perform one forward pass on a test image. Lower is better for efficiency. \\

\textbf{Para. Count} $\downarrow$ & Total number of trainable parameters. Lower indicates a more compact model. \\

\midrule
\multicolumn{2}{l}{\textit{Downstream Task Evaluation Metrics}} \\
\midrule
\textbf{Precision} $\uparrow$ & Proportion of correctly predicted positive samples. \\
\textbf{Recall} $\uparrow$ & Proportion of actual positives that are correctly identified. \\
\textbf{F1 Score} $\uparrow$ & Harmonic mean of Precision and Recall. \\
\textbf{IoU} $\uparrow$ & Intersection over Union between predicted and ground-truth regions. \\

\bottomrule
\end{tabular}
\end{table}

\subsection{Crop Type Mapping Method}

\subsubsection{XGBoost}

\begin{figure*}[h]
  \centering
  \includegraphics[width=\linewidth]{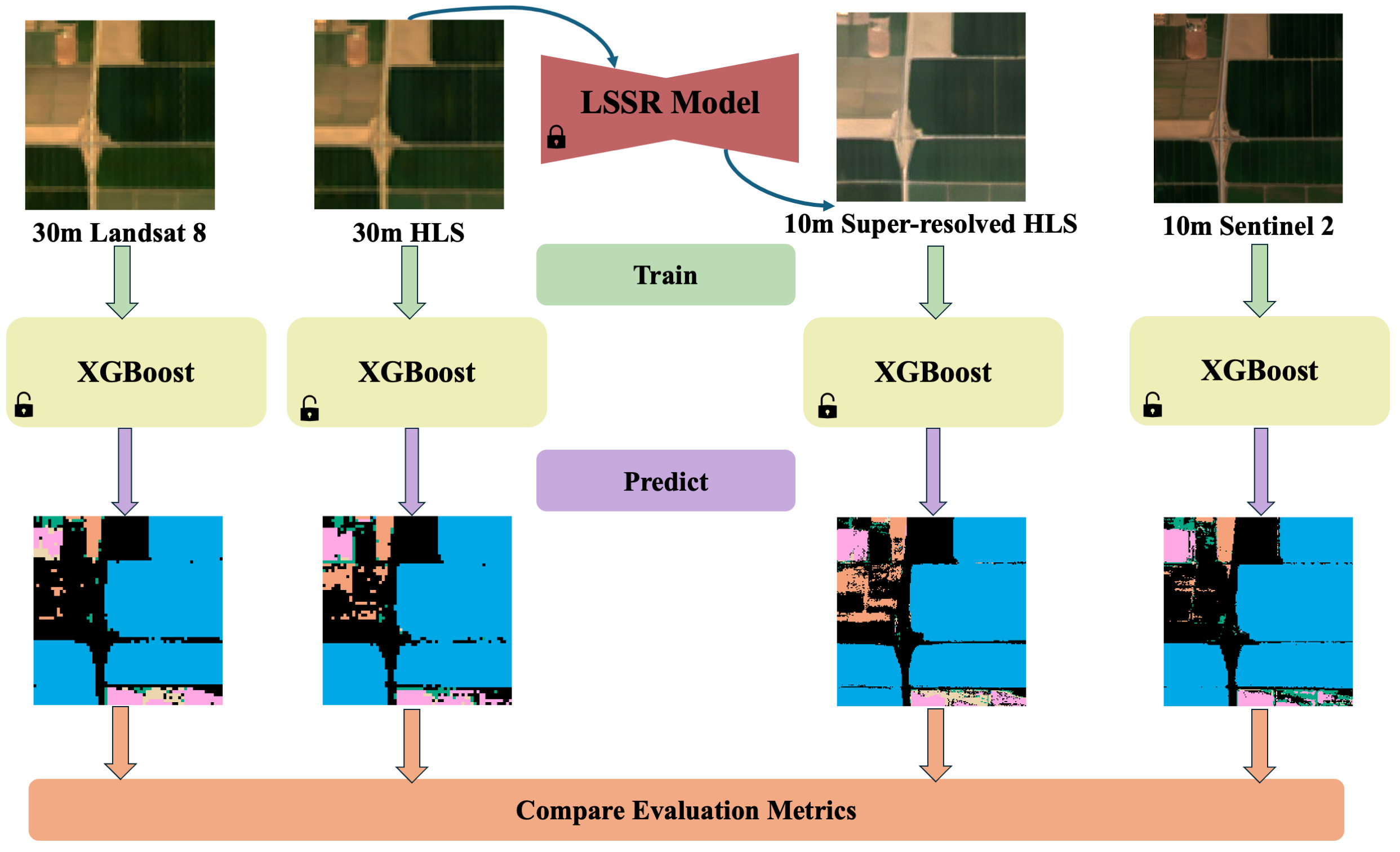}
  \caption{Crop Type Mapping Process.}
  \label{fig:mapping_process}
\end{figure*}

XGBoost \cite{chen2016xgboost} is a gradient boosting framework that sequentially adds weak learners to minimize a regularized objective function, balancing prediction accuracy and model complexity. Figure \ref{fig:mapping_process} shows the overall workflow for crop type mapping using different input resolutions and sensors. The goal is to evaluate the impact of super-resolved imagery on downstream classification accuracy. Specifically, we train and evaluate XGBoost classifiers using multiple image sources, including 30 m Landsat-8, 30 m HLS, and 10 m Sentinel-2. The proposed LSSR model generates 10 m super-resolved HLS images, which are further used for crop classification. We compare all predictions against the Sentinel-2-based classification map using standard accuracy metrics to validate the benefits of super-resolution for crop mapping.

\subsubsection{Evaluation Metrics}

We evaluate crop type classification performance using the following standard metrics, as shown in Table \ref{tab:metrics}. All metrics are computed per class and then averaged to assess overall classification performance across different spatial resolutions and data sources.

\subsection{Experiment settings}

The proposed LSSR architectures are implemented in PyTorch \cite{paszke2019pytorch}. The LR image pairs are 64 × 64 pixels, while HR pairs are 192 × 192 pixels. During LSSR training, AdamW optimizer is employed with an initial learning rate of $5e^{-5}$ scheduled using Constant. The batch size is set to 1 due to the hardware constraints. For both training and inference, we used an NVIDIA TITAN RTX GPU featuring 24GB of memory, with CUDA 12.2.

For the downstream crop type mapping task evaluation, experimental settings remain the same for all image resolutions and sensors. To address class imbalance, we compute class-specific weights based on the inverse frequency of class occurrences and assign a weight to each training sample accordingly. These sample weights are used to construct the XGBoost DMatrix for training. The XGBoost classifier is trained with a maximum depth of 20, a learning rate of 0.05, 128 histogram bins, and a subsample ratio of 0.7. We use histogram-based tree construction with GPU acceleration. 

\section{Results}
\label{sec:results}

This section reports the quantitative and qualitative results of our proposed LSSR method. 

\subsection{Overall Performance}

\begin{table}[htbp]
\centering
\scriptsize
\caption{SR Model Performance and Efficiency Comparison on LSSR Dataset. $\uparrow$: higher is better, $\downarrow$: lower is better. \textbf{Boldface}: best, \underline{Underlined}: second place. }
\label{tab:sr_perform}
\begin{tabular}{lccccccc}
\toprule
\textbf{Metric} & \textbf{Bicubic} & \textbf{SRCNN} & \textbf{EDSR} & \textbf{ESRGAN} & \textbf{StableSR} & \textbf{PiSA-SR} & \textbf{LSSR} \\
\midrule
\multicolumn{8}{l}{\textbf{RGB metrics}} \\
\midrule
PSNR $\uparrow$  & 20.03 & 23.84 & 29.75 & 29.28 & \underline{30.64} & 29.0466 & \textbf{32.63} \\
SSIM $\uparrow$  & 0.76 & \underline{0.79} & 0.78 & 0.72 & 0.78 & 0.77 & \textbf{0.84} \\
LPIPS $\downarrow$ & 0.28 & 0.28 & 0.28 & \underline{0.23} & \textbf{0.19} & 0.29 & 0.24 \\
FCL $\downarrow$  & 0.03 & 0.03 & 0.03 & \underline{0.02} & 0.03 & 0.02 & \textbf{0.01} \\
\midrule
\multicolumn{8}{l}{\textbf{IR metrics}} \\
\midrule
PSNR $\uparrow$  & 18.30 & 20.28 & \underline{21.26} & 19.70 & 19.40 & 19.46 & \textbf{23.99} \\
SSIM $\uparrow$  & 0.71 & \underline{0.73} & 0.70 & 0.69 & 0.71 & 0.68 & \textbf{0.78} \\
LPIPS $\downarrow$ & 0.37 & 0.50 & 0.41 & 0.30 & \textbf{0.31} & 0.46 & \underline{0.32} \\
FCL $\downarrow$   & 0.04 & \underline{0.02} & 0.04 & 0.03 & 0.04 & 0.03 & \textbf{0.01} \\
\midrule
\multicolumn{8}{l}{\textbf{Overall metrics}} \\
\midrule
SAM $\downarrow$   & 6.15 & 3.86 & 5.47 & \textbf{2.18} & 6.43 & 5.85 & \underline{3.79} \\
NDVI MSE $\downarrow$   & 0.08 & 0.11 & 0.15 & 0.09 & \underline{0.06} & 0.06 & \textbf{0.04} \\
Inference (sec) $\downarrow$ & 0.01 & 0.01 & 0.01 & 0.13 & 0.11 & 0.19 & 0.39 \\
Param Count $\downarrow$     & 0 & 57K & 40.73M & 12.70M & 1.56B & 1.29B & 1.29B \\
\bottomrule
\end{tabular}
\end{table}

\begin{figure}[htbp]
    \centering
    % --- 1 ---
    \begin{subfigure}{\textwidth}
        \centering
        \includegraphics[width=\linewidth]{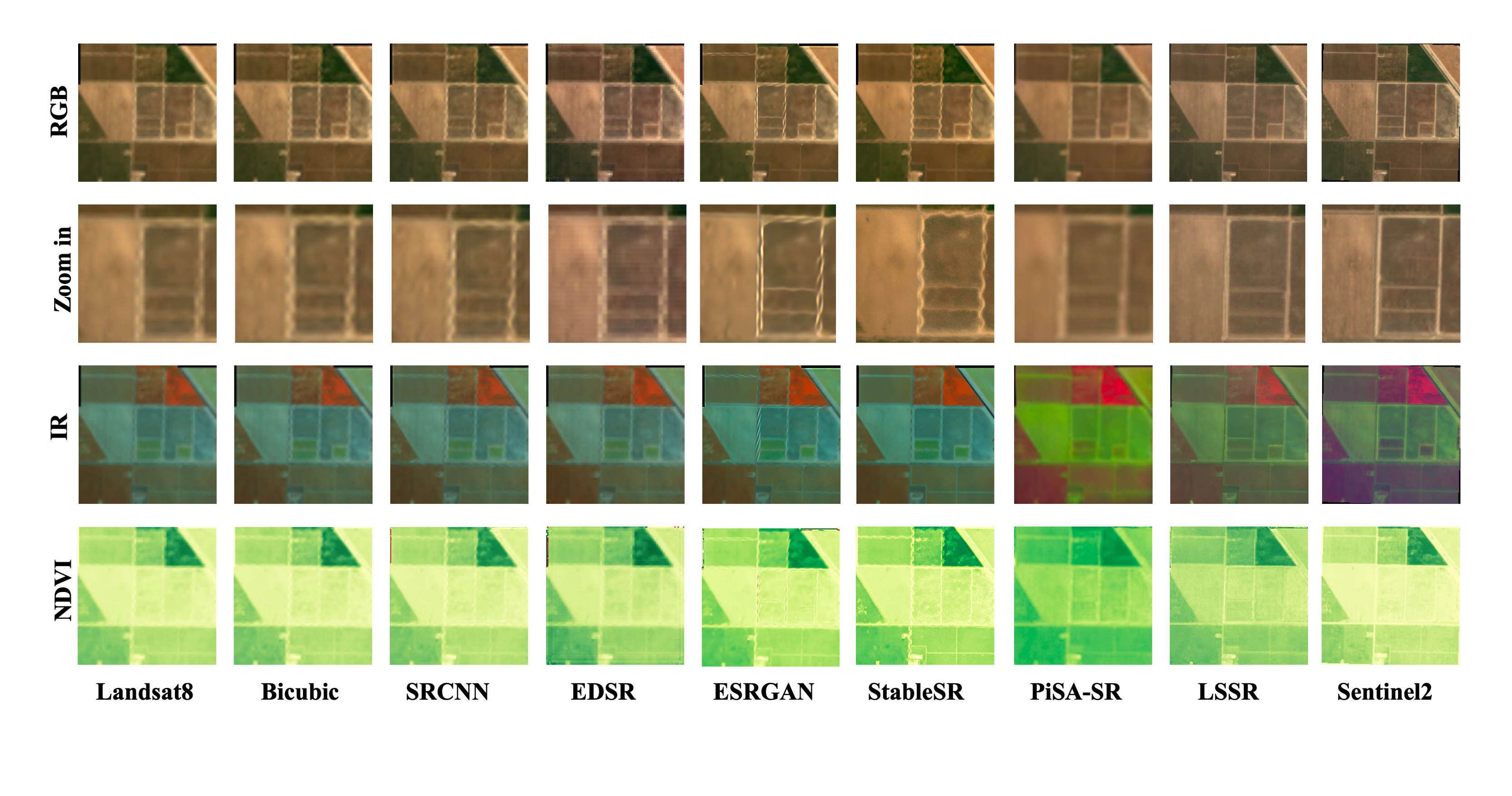}
        \caption{Sample 1.}
        \label{fig:sr_result_1}
    \end{subfigure}

    \vspace{1.5em} 

    % --- 2 ---
    \begin{subfigure}{\textwidth}
        \centering
        \includegraphics[width=\linewidth]{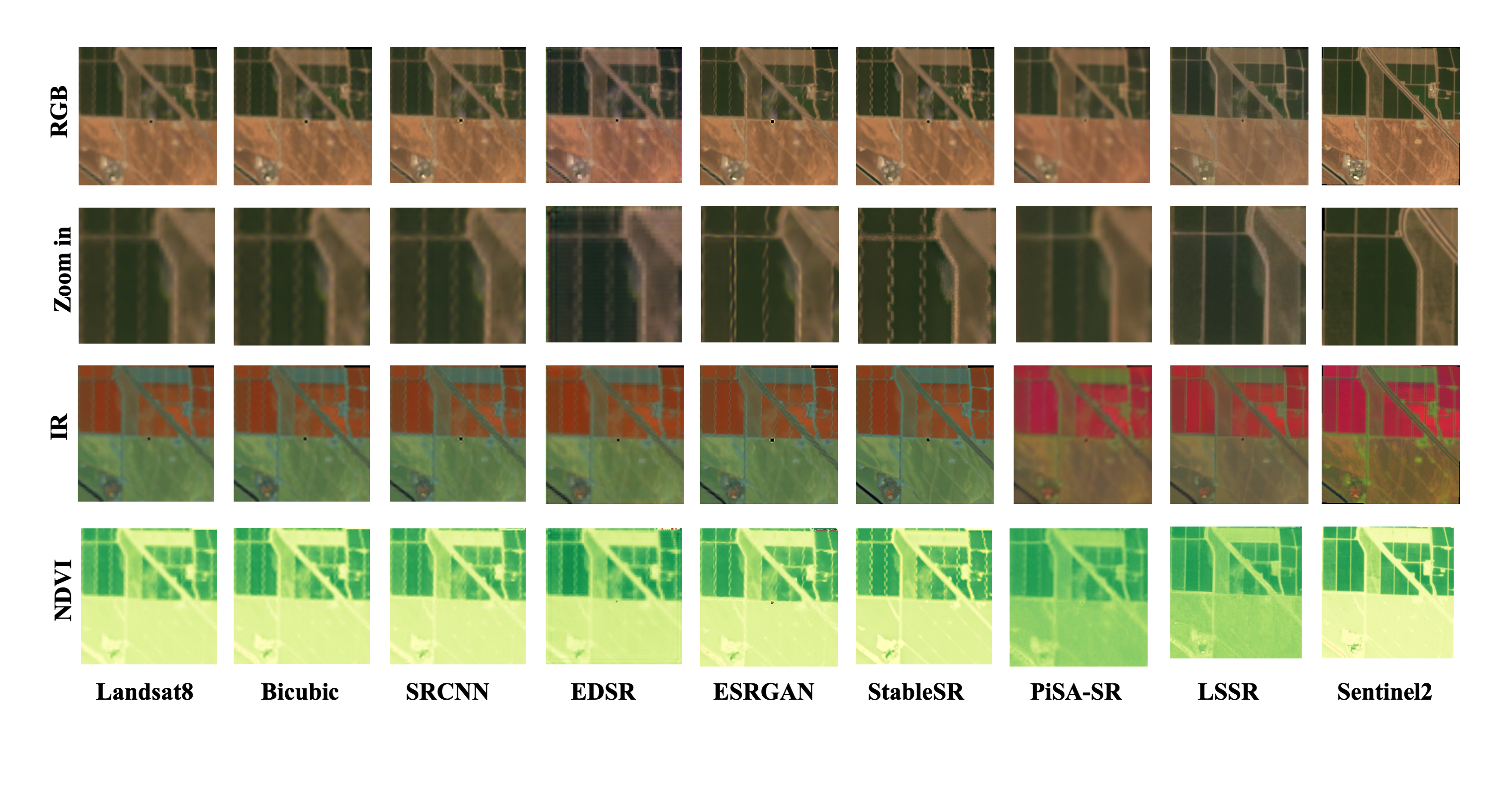}
        \caption{Sample 2.}
        \label{fig:sr_result_2}
    \end{subfigure}

    \caption{LSSR Model Result Samples. RGB composite (top row), Zoom in regions (row 2), IR composite (row 3), and NDVI visualization (bottom row).}
    \label{fig:sr_result}
\end{figure}

\begin{figure}[htbp]
    \centering
    % --- 3 ---
    \begin{subfigure}{\textwidth}
        \centering
        \includegraphics[width=\linewidth]{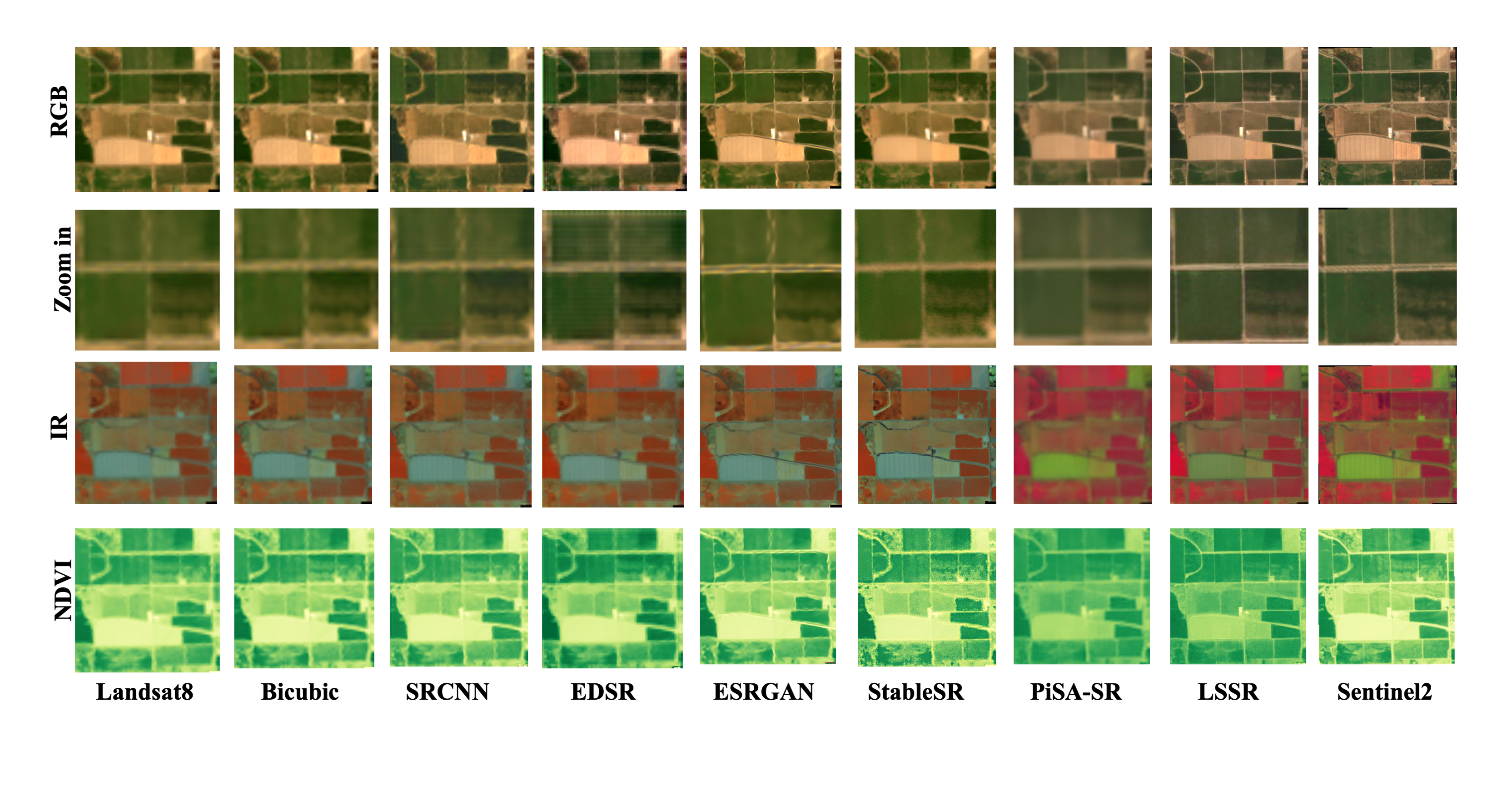}
        \caption{Sample 1.}
        \label{fig:sr_result_3}
    \end{subfigure}

    \vspace{1.5em}

    % --- 4 ---
    \begin{subfigure}{\textwidth}
        \centering
        \includegraphics[width=\linewidth]{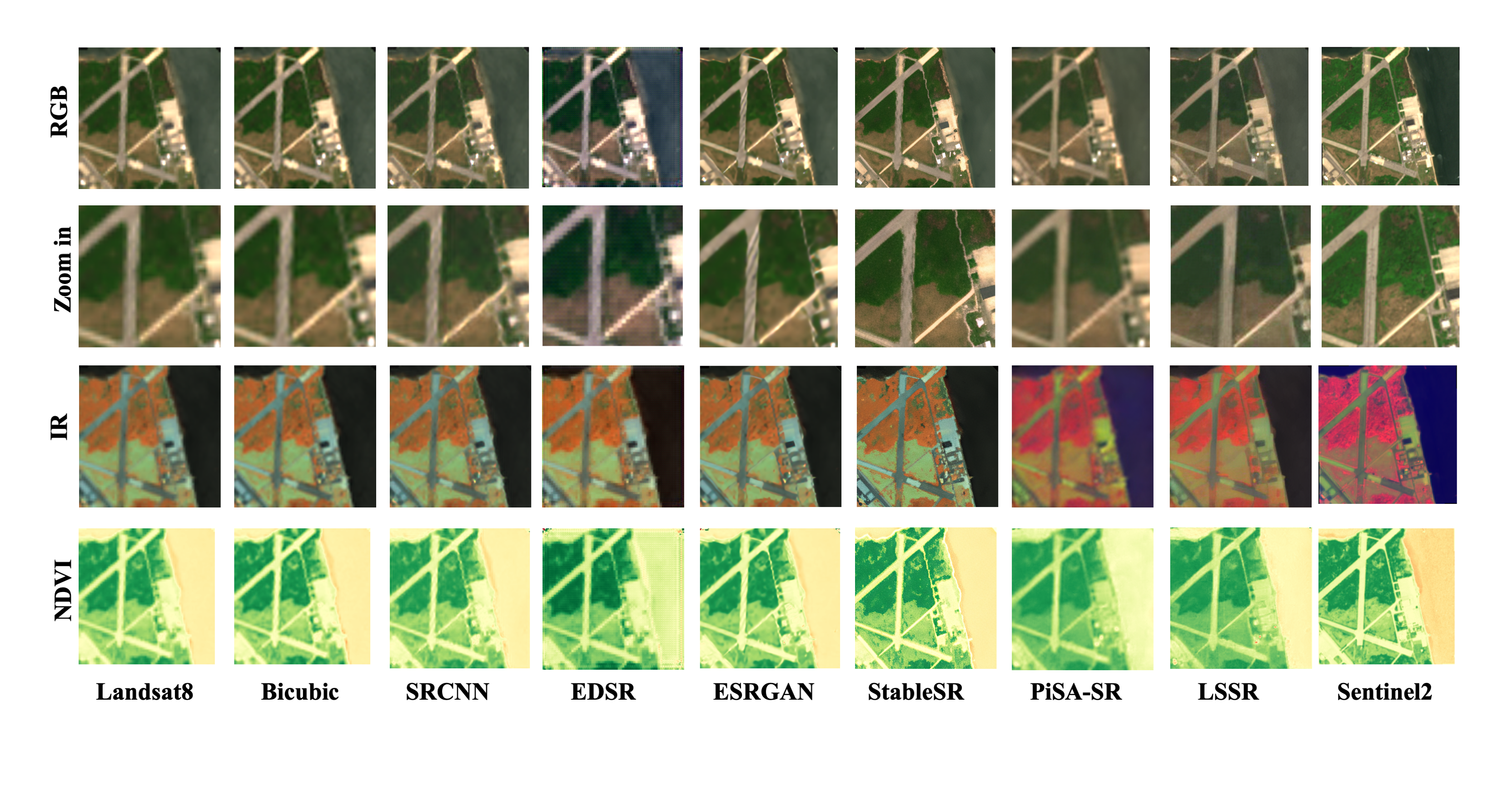}
        \caption{Sample 2.}
        \label{fig:sr_result_4}
    \end{subfigure}

    \caption{More LSSR Model Result Samples. RGB composite (top row), Zoom in regions (row 2), IR composite (row 3), and NDVI visualization (bottom row).}
    \label{fig:sr_result_continue}
\end{figure}

As shown in Table \ref{tab:sr_perform}, our proposed method LSSR achieves the best overall performance across both RGB and IR bands, indicating strong reconstruction fidelity and consistency. For RGB metrics, LSSR obtains the highest PSNR (32.63) and SSIM (0.84), indicating excellent perceptual and structural fidelity. It also achieves the lowest FCL (0.01), highlighting strong feature consistency, although LPIPS is slightly higher than StableSR. For IR metrics, LSSR significantly outperforms all baselines, especially in PSNR (23.99) and SSIM (0.78), confirming its effectiveness in enhancing low-quality infrared inputs. However, LSSR does not outperform GAN-based approaches such as ESRGAN and StableSR in LPIPS, indication that its reconstructions may appear less visually sharp. Regarding NDVI MSE, a cross-spectral evaluation metric reflecting vegetation index accuracy, LSSR achieves the lowest error (0.04), demonstrating superior semantic consistency across spectral bands. Overall, LSSR prioritizes accuracy and scientific utility over other methods.

Figure \ref{fig:sr_result} and \ref{fig:sr_result_continue} presents the visual comparison of different SR methods on agricultural scenes. Compared with baseline methods (e.g., Bicubic, SRCNN, EDSR, ESRGAN, StableSR, and PISA-SR), our proposed LSSR produces visually sharper and realistic crop field boundaries, better texture restoration, and more accurate spectral consistency. In the RGB composite, LSSR restores fine-grained field structures with enhanced clarity, closely resembling the high-resolution Sentinel-2 reference. Notably, traditional models like ESRGAN introduce checkerboard artifacts, while StableSR, despite producing cleaner edges, fails to preserve subtle contrast differences between adjacent fields.

In the Zoom in row, the visual differences among models become more apparent. Bicubic and SRCNN produce blurry textures with little structural detail preserved. EDSR and ESRGAN enhance sharpness but often introduce unnatural artifacts and edges. StableSR generates relatively clear boundaries but suffers from oversmoothing. PiSA-SR shows moderate improvement yet still loses fine structures. By contrast, LSSR reconstructs sharper field boundaries and more consistent textures, yielding results that are visually closer to the Sentinel-2 reference.

The IR composite results highlight LSSR’s strength in preserving spectral fidelity. While PiSA-SR introduces noticeable color distortions (e.g., excessive green or red tint), LSSR maintains a more natural tone and band alignment, reducing false colors and improving semantic consistency. 

Moreover, Figure \ref{fig:sr_result_4} compares multiple super-resolution methods applied to an agricultural–coastal area with roads and vegetation. The Bicubic, SRCNN, EDSR exhibit substantial spatial blurring. ESRGAN and StableSR reconstruct road textures, yet over-sharpen road edges and have perceptual artifacts. In contrast, LSSR generates natural details that closely match the Sentinel-2 reference.

SAM metric in Table 4 shows that ESRGAN achieves the lowest spectral distortion (2.18), outperforming all competing methods. Although LSSR attains a relatively small SAM value (3.79), it exhibits better spatial reconstruction in both PSNR and SSIM. In contrast, StableSR and PiSA-SR, which are recent diffusion-based and semantic-guided SR models, yield higher SAM scores (6.43 and 5.85, respectively), indicating larger spectral deviations due to their knowledge priors from natural images.

Finally, in terms of NDVI, which reflects vegetation distribution and health, LSSR generates a more accurate gradient, minimizing noise in low-contrast areas and overexposed zones. It exhibits superior alignment with the Sentinel-2 reference, especially along field boundaries and heterogeneous patches, validating its effectiveness in cross-modal reconstruction. Overall, LSSR consistently provides visually and semantically realistic outputs, confirming the quantitative improvements shown in Table \ref{tab:sr_perform}.

On the other hand, in terms of inference efficiency, as shown in Table \ref{tab:sr_perform}, while LSSR requires longer inference time (0.3915 sec) than previous models like SRCNN, it maintains a manageable parameter size (1.29B), comparable to PiSA-SR. Overall, LSSR offers the best trade-off between reconstruction accuracy and cross-spectral consistency, particularly excelling in the challenging infrared and NDVI domains.

\subsection{Ablation Study}
\label{sec:ablation}

\begin{table}[htbp]
\centering
\scriptsize
\caption{Ablation Study of the Proposed LSSR. $\uparrow$: higher is better, $\downarrow$: lower is better. \textbf{Boldface}: best.}
\label{tab:ablation}
\begin{tabular}{lccccc}
\toprule

\textbf{Metric} & 
\textbf{Plain} & 
\makecell[c]{\textbf{+DEM LC} \\ \textbf{encoders}} & 
\makecell[c]{\textbf{+Temporal} \\ \textbf{encoder}} & 
\makecell[c]{\textbf{+Cross} \\ \textbf{attention}} & 
\makecell[c]{\textbf{+$\mathrm{loss}_{\mathrm{fft}}$}} \\

\midrule
\multicolumn{6}{l}{\textbf{RGB metrics}} \\
\midrule
PSNR $\uparrow$  & 29.04 & 32.23 & 32.25 & 32.32 & 32.35 \\
SSIM $\uparrow$  & 0.77 & 0.81 & 0.80 & 0.81 & 0.82 \\
LPIPS $\downarrow$ & 0.29 & 0.30 & 0.30 & 0.29 & 0.25 \\
FCL $\downarrow$   & 0.03 & 0.02 & 0.02 & 0.02 & 0.02 \\
\midrule
\multicolumn{6}{l}{\textbf{IR metrics}} \\
\midrule
PSNR $\uparrow$  & 19.46 & 19.19 & 19.21 & 19.22 & 19.29 \\
SSIM $\uparrow$  & 0.68 & 0.65 & 0.66 & 0.66 & 0.67 \\
LPIPS $\downarrow$ & 0.46 & 0.34 & 0.35 & 0.33 & 0.34 \\
FCL $\downarrow$   & 0.03 & 0.03 & 0.03 & 0.03 & 0.03 \\
\midrule
\multicolumn{6}{l}{\textbf{Overall metrics}} \\
\midrule
NDVI MSE $\downarrow$   & 0.06 & 0.05 & 0.04 & 0.04 & 0.04 \\
Infer. (sec) $\downarrow$ & 0.19 & 0.37 & 0.37 & 0.37 & 0.37 \\
Param. $\downarrow$     & 1.29B  & +0.052M & +0.052M & +0.317M & +0 \\

\midrule

\textbf{Metric} & 
\makecell[c]{\textbf{+10x} \\ $\mathrm{loss}_{\mathrm{ndvi}}$ \\ (not used)} & 
\makecell[c]{\textbf{+20x} \\ $\mathrm{loss}_{\mathrm{ndvi}}$} & 
\makecell[c]{\textbf{+30x} \\ $\mathrm{loss}_{\mathrm{ndvi}}$ \\ (not used)} &
\makecell[c]{\textbf{IR specific} \\ \textbf{LoRA} \\ (not used)} &
\makecell[c]{\textbf{SAR-Guided} \\ \textbf{Fusion}} \\

\midrule
\multicolumn{6}{l}{\textbf{RGB metrics}} \\
\midrule
PSNR $\uparrow$  & 32.45 & 32.46 & 32.41 & 32.22 & \textbf{32.63} \\
SSIM $\uparrow$  & 0.83 & 0.83 & 0.83 & 0.82 & \textbf{0.84} \\
LPIPS $\downarrow$ & 0.25 & 0.25 & 0.25 & 0.27 & \textbf{0.24} \\
FCL $\downarrow$   & 0.02 & 0.02 & 0.02 & 0.03 & \textbf{0.01} \\
\midrule
\multicolumn{6}{l}{\textbf{IR metrics}} \\
\midrule
PSNR $\uparrow$  & 19.36 & 23.55 & 23.12 & 22.34 & \textbf{23.99} \\
SSIM $\uparrow$  & 0.68 & \textbf{0.78} & 0.77 & 0.73 & \textbf{0.78} \\
LPIPS $\downarrow$ & 0.41 & \textbf{0.32} & 0.33 & 0.35 & 0.32 \\
FCL $\downarrow$   & 0.02 & 0.02 & 0.02 & 0.02 & \textbf{0.01} \\
\midrule
\multicolumn{6}{l}{\textbf{Overall metrics}} \\
\midrule
NDVI MSE $\downarrow$   & 0.04 & 0.04 & 0.04 & 0.06 & \textbf{0.04} \\
Infer. (sec) $\downarrow$ & 0.3915 & 0.3915 & 0.3915 & 0.4065 & 0.3985 \\
Param. $\downarrow$     & +0 & +0 & +0 & +4.056M & +0.280 M \\

\bottomrule
\end{tabular}
\end{table}

Table \ref{tab:ablation} presents the ablation results of the proposed LSSR model by progressively integrating different components and supervision strategies. Starting from a plain model, we evaluate the contribution of each module on RGB/IR quality and cross-spectral consistency.

\textbf{Knowledge Encoder Contributions.}
Adding DEM and land cover (LC) encoders leads to a substantial improvement across RGB and IR metrics, especially PSNR (+3.2dB for RGB, +0.7dB for IR) and NDVI MSE (from 0.06↓ to 0.05↓), demonstrating the effectiveness of auxiliary spatial information. Incorporating a temporal encoder yields marginal gains, while cross-attention further enhances performance, reducing RGB FCL to 0.02 and improving SSIM to 0.81.

\textbf{Knowledge Constraint Attention Mechanism.}
The introduction of cross attention, which leverages DEM, LC, and temporal encoder embeddings as keys and values, substantially increases the trainable parameters from $+0.052M$ (temporal encoder) to $+0.317M$. As a result, the corresponding performance improvements are also significant. Specifically, for RGB metrics, PSNR increases only from 32.25 to 32.32 and SSIM from 0.80 to 0.81. However, IR metrics show similarly limited gains.

\textbf{Spectral-aware Loss Terms.}
The introduction of frequency-domain loss (fft loss) and NDVI-guided supervision progressively refines the model outputs. Notably, +10× ndvi loss already brings strong gains in RGB SSIM (0.83) and FCL (0.01), and pushing to +20× ndvi loss further improves IR PSNR to 24.55 and reduces NDVI MSE to 0.04, the best among all variants. The +30× version slightly saturates or regresses in performance, suggesting over-regularization. Finally, we chose 20× ndvi loss into the final LSSR model architecture.

\textbf{IR-specific LoRA.}
Introducing an IR-specific LoRA branch results in degraded RGB and IR quality (e.g., RGB PSNR drops to 32.22, IR SSIM to 0.73), while increasing parameter count significantly (+4.056M).

\textbf{SAR-Guided Fusion.}
The SAR-guided fusion module introduces cross-modal interactions by explicitly incorporating VH and VV features to guide the reconstruction of RGB/IR bands. Adding only $+0.280$M parameters is significantly lower than the $+4.056$M required by the IR-specific LoRA, but the performance improvements are substantial and consistent across both RGB and IR metrics. For RGB, PSNR improves to 32.63 and SSIM to 0.84, while perceptual metrics such as LPIPS and FCL achieve their best values (0.24 and 0.01, respectively). For IR, SAR guidance leads to the highest PSNR (23.99) and competitive SSIM (0.78), indicating a strong enhancement in structural fidelity. Moreover, the overall NDVI MSE is reduced to 0.04, further confirming the spectral accuracy with the SAR integration.

\section{Application: Crop Type Mapping using Super-resolved HLS}
\label{sec:application}

\subsection{Overall Performance}

\begin{table}[htbp]
\centering
\scriptsize
\renewcommand{\arraystretch}{1.2}
\setlength{\tabcolsep}{2.8pt}
\caption{XGBoost classification results across different resolutions.}
\label{tab:crop_mapping}
\begin{tabular}{l|cccc|cccc|cccc|cccc}
\toprule
\textbf{ } & \multicolumn{4}{c|}{\textbf{30m Landsat 8}} & \multicolumn{4}{c|}{\textbf{30m HLS}} & \multicolumn{4}{c|}{\textbf{10m super-resolved HLS}} & \multicolumn{4}{c}{\textbf{10m Sentinel 2}} \\
& Prec. & Recall & F1 & IoU & Prec. & Recall & F1 & IoU & Prec. & Recall & F1 & IoU & Prec. & Recall & F1 & IoU \\
\hline
Background & 0.84 & 0.88 & 0.86 & 0.76 & 0.86 & \textbf{0.93} & \textbf{0.89} & 0.81 & \textbf{0.87} & 0.92 & \textbf{0.89} & 0.81 & 0.87 & 0.91 & \textbf{0.89} & 0.80 \\
Corn       & 0.89 & \textbf{0.85} & 0.87 & 0.77 & 0.90 & 0.83 & 0.87 & 0.76 & \textbf{0.93} & 0.85 & \textbf{0.89} & 0.81 & 0.91 & 0.82 & 0.87 & 0.76 \\
Soybean    & \textbf{0.80} & 0.82 & 0.81 & 0.68 & 0.77 & 0.82 & 0.79 & 0.66 & 0.78 & \textbf{0.89} & \textbf{0.81} & \textbf{0.70} & 0.76 & 0.85 & 0.80 & 0.67 \\
Overall  & 0.84 & 0.85 & 0.85 & 0.73 & 0.84 & 0.86 & 0.85 & 0.74 & \textbf{0.86} & \textbf{0.87} & \textbf{0.86} & 0.77 & 0.84 & 0.86 & 0.85 & 0.74 \\
% Weighted avg & 0.85 & 0.85 & 0.85 & -- & 0.86 & \textbf{0.86} & \textbf{0.86} & -- & \textbf{0.87} & \textbf{0.86} & \textbf{0.86} & -- & 0.86 & 0.86 & 0.86 & -- \\
\bottomrule
\end{tabular}
\end{table}

\begin{figure}[htbp]
    \centering

    \begin{subfigure}{\textwidth}
        \centering
        \includegraphics[width=0.9\linewidth]{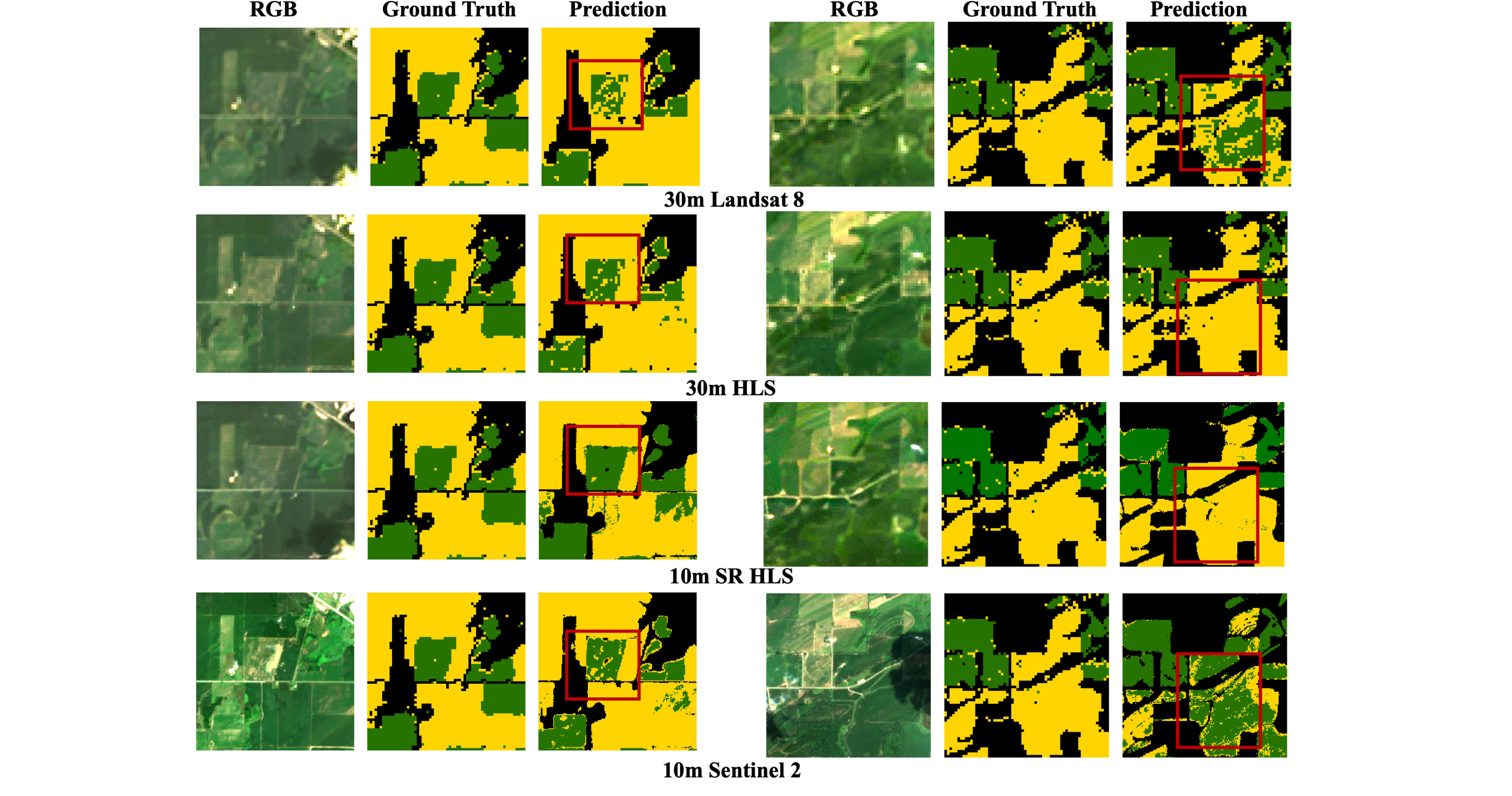}
        \caption{Sample 1.}
        \label{fig:cp_result_1}
    \end{subfigure}

    \begin{subfigure}{\textwidth}
        \centering
        \includegraphics[width=0.9\linewidth]{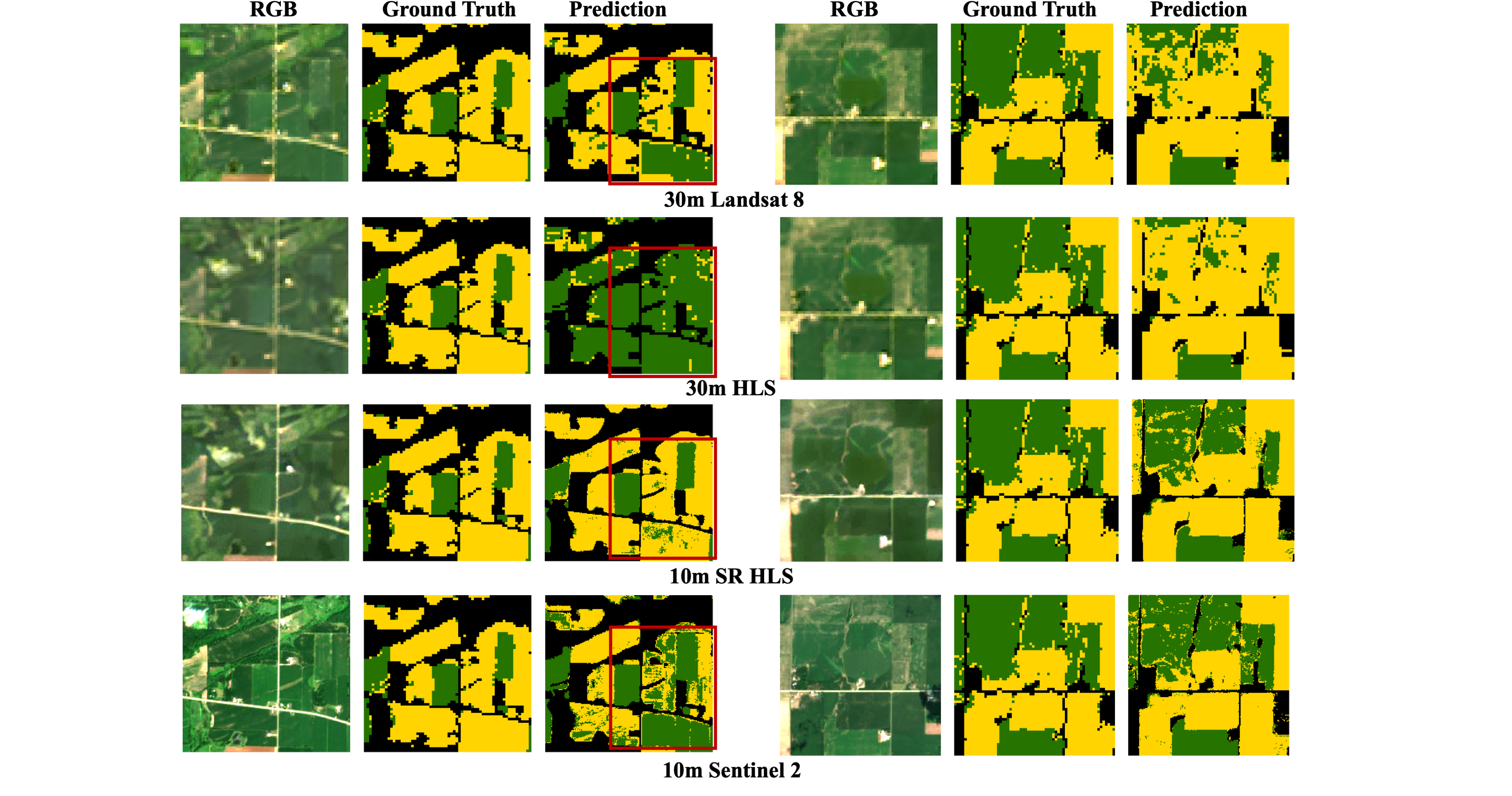}
        \caption{Sample 2.}
        \label{fig:cp_result_2}
    \end{subfigure}

    \begin{subfigure}{\textwidth}
        \centering
        \includegraphics[width=0.9\linewidth]{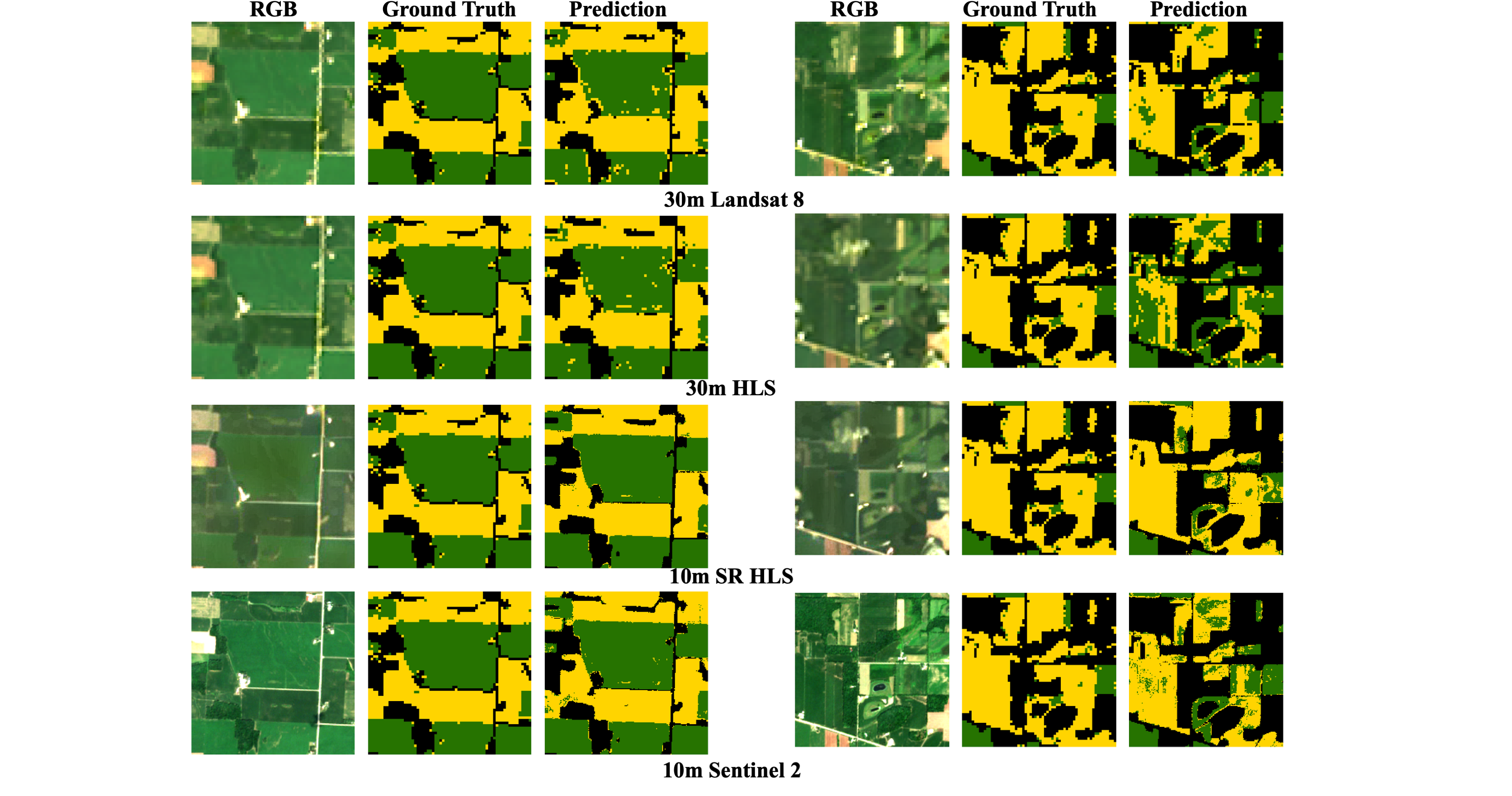}
        \caption{Sample 3.}
        \label{fig:cp_result_3}
    \end{subfigure}

    \caption{XGBoost Prediction Result Samples.}
    \label{fig:cp_result}
\end{figure}

Table \ref{tab:crop_mapping} compares the XGBoost classification performance across different input resolutions: 30 m Landsat 8, 30 m HLS, 10 m super-resolved HLS, and 10 m Sentinel-2. Each configuration is evaluated on four key metrics: precision, recall, F1-score, and IoU. The 30 m HLS baseline performs well, with a macro F1-score of 0.85 and recall reaching 0.93 for the background class. However, it exhibits slightly lower performance on soybean (F1 = 0.79, IoU = 0.66), indicating limitations in classifying spectrally similar crops at coarse resolution. By contrast, the 10 m super-resolved HLS by LSSR consistently improves performance across all classes. It achieves the highest corn F1-score (0.89), and its macro average metrics (Precision = 0.86, Recall = 0.87, F1 = 0.86, IoU = 0.77) match or exceed those of Sentinel-2. Importantly, super-resolved HLS narrows the gap with native 10 m Sentinel-2, validating the effectiveness of resolution enhancement for downstream classification tasks. Soybean classification also benefits from SR: its IoU improves from 0.66 (HLS) to 0.70, and F1 remains stable at 0.81. Compared to Sentinel-2, the SR-HLS results are competitive, with only marginal differences across all categories, suggesting the model’s potential as a practical alternative when 10 m observations are unavailable.

Figure \ref{fig:cp_result_1}, Figure \ref{fig:cp_result_2}, and Figure \ref{fig:cp_result_3} showcase side-by-side visual comparisons of classification results across different input sources: 30 m Landsat 8, 30 m HLS, 10 m super-resolved HLS, and 10 m Sentinel-2. Each row includes the RGB image, ground truth, and prediction. Three geographically distinct regions are shown to demonstrate generalization. Across all three figures, the 30 m Landsat 8 and 30 m HLS inputs consistently produce over-smoothed or fragmented classification maps. Notably, the boundaries between corn (yellow) and soybean (green) are poorly delineated in these baselines, with Landsat 8 exhibiting the most significant confusion and HLS showing slight improvement.

The 10 m super-resolved HLS results show clear improvement over their 30 m counterparts. Field boundaries become more distinguishable, and predictions better match the ground truth structure, particularly in complex or mixed-pixel regions (e.g., red boxes in Figures \ref{fig:cp_result_1} and \ref{fig:cp_result_2}). Compared with native Sentinel-2, SR-HLS performs comparably in most regions, with only minor artifacts or omissions near object edges. In Figure \ref{fig:cp_result_3}, which includes a particularly heterogeneous landscape, the benefit of super-resolution is especially prominent. The 30 m inputs fail to capture narrow strips of soybean fields, while both SR-HLS and Sentinel-2 recover them well. Moreover, SR-HLS maintains semantic coherence even in visually ambiguous zones (e.g., shaded or noisy regions in RGB). Overall, these visualizations demonstrate that our super-resolved HLS enhances spatial precision and semantic consistency, bridging the gap between LR observations and HR Sentinel-2 references. It is worth noting that the goal of this physically constrained framework is to achieve physically consistent reconstructions that align with native high-resolution observations. From this perspective, the comparable performance to Sentinel-2 validates that the super-resolved HLS preserves the underlying radiometric and structural integrity of the original data, which is more critical for physically meaningful downstream analysis.

\section{Discussion}
\label{sec:discussion}

\subsection{Interpretation and Analysis}

\subsubsection{Visual Artifacts}

In comparative experiments (Figure \ref{fig:sr_result}-\ref{fig:sr_result_continue}), we observed that some previous models, such as ESRGAN and StableSR, generated curved or distorted line artifacts, especially in field and road boundaries. These artifacts can be attributed to the models’ mechanisms. Specifically, adversarial training in GAN-based methods (e.g., ESRGAN) overemphasized high-frequency details to improve image sharpness. However, in RS images, sharpness can distort geometrically regular structures. Previous experiments comparing SR methods in RS images also show similar results \cite{wang2023review, gong2021enlighten}.

Similarly, earlier diffusion-based models such as StableSR are primarily trained on natural image datasets that contain irregular object textures (e.g., human faces, natural scenes). When applied to RSSR data with highly structured patterns and clear linear features, they can deform spatial geometry, and hallucinate distorted textures \cite{li2025diffusion}.

In contrast, our proposed LSSR model incorporates physically constrained attention modules and NDVI-guided consistency terms, which jointly enforce spectral fidelity and geometric stability. This design effectively suppresses visually pleasing yet physically implausible textures, preserving the linear and spatial continuity of agricultural field boundaries. The results highlight the importance of incorporating domain-specific physical constraints when adapting generative restoration models to remote sensing applications.

\subsubsection{Effectiveness of Components}

The ablation study in Table \ref{tab:ablation} highlights the effectiveness of each component in the proposed LSSR framework. First, starting from a plain backbone, the inclusion of DEM, land cover, and temporal information introduces valuable spatial priors, significantly improving both RGB and IR performance. The addition of cross-attention further enhances feature integration, improving boundary quality and semantic alignment. Second, among all modifications, the incorporation of spectral-aware supervision, which is the NDVI-guided loss, plays a central role in improving spectral consistency. Scaling the NDVI loss from 10× to 20× leads to measurable gains in IR and NDVI MSE metrics. This suggests that while NDVI supervision is beneficial in optimization signals across spectral bands.

Notably, the introduction of an IR-specific LoRA branch increases model complexity (+4M parameters) but degrades performance in both RGB and IR outputs. This implies that excessive modality decoupling may harm the shared spectral representation, underscoring the importance of joint modeling over hard separation in multi-spectral reconstruction tasks.

Texture guided RSSR has been popular in recent years. For example, a saliency map is a visual representation that highlights the most important or attention-worthy regions in an image, which can reflect texture complexity and guide the generator in restoring regions with varying levels of detail, such as SD-GAN \cite{ma2019sd} and Saliency-Driven Feedback GAN SDFBGAN \cite{wu2020remote}. On the other hand, the incorporation of SAR images provides additional prior information for RGB/IR reconstruction because SAR can penetrate the clouds and reflect structural information of the surface \cite{shu2025restoredit}. Our results also show that SAR-guided fusion offers the most effective trade-off between model efficiency (0.39 sec/image) and performance (highest PSNR/SSIM: 32.63/0.84, and lowest NDVI MSE: 0.04).

The qualitative results in Figure \ref{fig:cp_result_1}, Figure \ref{fig:cp_result_2}, and Figure \ref{fig:cp_result_3} illustrate the clear benefits of applying LSSR to medium-resolution HLS imagery for crop classification. Compared to the 30 m inputs from Landsat 8 and native HLS, the 10 m super-resolved HLS enhances spatial detail, producing smoother and more coherent classification maps that better align with high-resolution Sentinel-2 references. Notably, SR-HLS improves the delineation of narrow field boundaries and mixed-pixel regions, which are often misclassified or oversmoothed at 30 m resolution. The task evaluation of S2DR3 model \cite{akhtman2024sentinel2} also shows that S2 and S2DR3 were very similar on crop type mapping classification, confirming the significant potential of S2DR3 for high-resolution crop mapping \cite{chanev2025evaluation}.

In all three geographic regions, the super-resolved predictions preserve crop shapes and boundaries more faithfully, recovering small-scale structures such as thin soybean strips or irregular field edges that are absent in the coarse-resolution results. This indicates that the learned super-resolution process not only improves image sharpness, but also retains semantically meaningful information relevant to the classification task.

Overall, these findings reinforce the potential of LSSR super-resolved products for downstream applications in agricultural monitoring, particularly in areas where HR satellite coverage is limited or inconsistent. The results also support the integration of super-resolution as a potential preprocessing step in RS classification pipelines.

\subsection{Limitations and Future Works}

Our proposed LSSR demonstrates strong performance across multiple metrics and datasets, supports downstream crop type mapping task evaluation, but there are still several areas that merit further exploration. First, the LSSR model partially relies on semantic guidance from Contrastive Language-Image Pre-training (Open CLIP model) \cite{cherti2023reproducible} text embeddings, which may be too generic to provide detailed information (e.g., "a crop field") in the context of agricultural RS. In particular, the lack of explicit image-text alignment feedback during training may lead to semantic misalignment going unnoticed. For instance, in Figure \ref{fig:loss_curves}, without the regularization effect of text embeddings, the CSD loss curve has obvious oscillations. To address this, future work could explore the integration of RSCLIP \cite{li2023rsclip} or AgriCLIP \cite{nawaz2024agriclip} models pretrained specifically on RS data or the incorporation of alignment-aware objectives.

Second, the model does not leverage crop-structural priors, which are known to improve performance in SR tasks. For agricultural imagery, crop row patterns or regular textures may be better maintained by explicitly modeling such priors. Crop structural parameters, such as leaf area index, stem height, stem density, and canopy gap fraction, are directly linked to plant geometry and biophysical status \cite{naidoo2022machine, cui2010using}. However, most SR models, including ours, neglect these biophysical cues, relying solely on image appearance. This can lead to over-smoothed textures or unnatural patterns, especially in structured fields where row planting and directional canopy orientation dominate. 

Third, experiments in this study were conducted on 64×64 and 192×192 patches to maintain training and computational efficiency on single GPU. The current implementation uses fixed-size inputs during inference. Future work will integrate and evaluate a tiling strategy to enable large-scale RSSR with limited memory usage.

Fourth, the LSSR model remains sensitive to cloud covers, which can obscure important spatial and spectral information in the input. RESTORE-DiT shows that diffusion models could also benefit cloud removal \cite{shu2025restoredit}. Future work may benefit from a unified framework that jointly addresses cloud removal, dehazing, SR, and more downstream classification and regression tasks \cite{qu2024mtlsc}, potentially via multi-task learning or sequential enhancement pipelines. 

Finally, this study specifically targets crop-dominated regions, as the goal of improving HLS data for agricultural monitoring guided the workflow, including data collection, model design, and task evaluation. However, validation on non-crop areas and other land cover regions was not included, a more general model architecture can be extended to other land-cover types.

\section{Conclusions}
\label{sec:conclusions}

In this study, we created a multi-modal RSSR dataset comprising paired 30 m Landsat-8 and 10 m Sentinel-2 images, and proposed an efficient LSSR framework for enhancing medium-resolution satellite imagery, with a particular focus on precision agriculture applications such as crop type classification. The proposed LSSR architecture is built on frozen pretrained Stable Diffusion, augmented with cross-modal attention mechanisms to incorporate auxiliary knowledge (DEM, land cover, month information) and SAR guidance (VH and VV images). It further integrates LoRA adapters and a tailored Fourier Transform and Vegetation Index loss to balance spatial detail and spectral fidelity.

Through extensive quantitative and qualitative evaluations, LSSR achieves superior overall performance in RSSR, particularly in delineating crop boundaries. It obtains the highest PSNR/SSIM scores on both RGB (32.63/0.84) and IR (23.99/0.78) reconstruction, while reducing NDVI MSE to 0.04 and maintaining efficient inference (0.39 s per image). We also demonstrate that the LSSR model can be effectively transferred to HLS super-resolution, where the super-resolved imagery yields more reliable crop classification results (F1: 0.86) compared to Sentinel-2 (F1: 0.85). Looking ahead, we highlight promising future directions in tailoring RS-specific and agriculture-specific text embeddings, incorporating crop-structural priors that link with plant geometry and biophysical status, and advancing toward unified low-level vision frameworks.

\section{Data Availability Statement}
\label{sec:data_avail}

The dataset used in this study is available in the Figshare repository at 
\href{https://doi.org/10.6084/m9.figshare.30062527.v3}{https://doi.org/10.6084/m9.figshare.30062527.v3} \cite{yang2025lssr}, licensed under CC-BY 4.0.

\appendix

\section{Pseudocode}
\label{sec:pseudo}
In this section, we provide the pseudocode of the proposed LSSR model in Algorithm \ref{alg:lssr}.

\begin{algorithm}[H]
\caption{LSSR}
\label{alg:lssr}
\begin{algorithmic}[1]
\Require RGB image $I_{rgb}$, IR image $I_{ir}$; DEM $D$, LandCover $L$, Month $m$; Sentinel-1 $S_1$; VAE $\mathcal{V}$, UNet $\mathcal{U}$; text prompts $\mathcal{P}$; scheduler time $t$
\Ensure Refined RGB $\hat{I}_{rgb}$, Refined IR $\hat{I}_{ir}$
\State \textbf{Encode to latents:} $z_{rgb}\!\leftarrow\!\mathcal{V}.\mathrm{enc}(I_{rgb})$, \ $z_{ir}\!\leftarrow\!\mathcal{V}.\mathrm{enc}(I_{ir})$
\State \textbf{Build knowledge features:} $\mathbf f_{DEM}\!\leftarrow\!\mathrm{Enc}_{dem}(D),\ \mathbf f_{LC}\!\leftarrow\!\mathrm{Enc}_{lc}(L),\ \mathbf f_{month}\!\leftarrow\!\mathrm{Enc}_{mon}(m)$
\State \textbf{Aggregate:} $\mathbf z_{aux}\!\leftarrow\!\mathbf f_{DEM}+\mathbf f_{LC}+\mathbf f_{month}$ \hfill (Eq.(5))
\State \textbf{Knowledge injection (Alg.\ref{alg:know_guieded}):}
       $z_{rgb}\!\leftarrow\!\mathrm{KnowInject}(z_{rgb}, \mathbf z_{aux})$, \ 
       $z_{ir}\!\leftarrow\!\mathrm{KnowInject}(z_{ir}, \mathbf z_{aux})$
\State \textbf{Text cond.:} $(\mathbf e_{pos},\mathbf e_{neg},\mathbf e_{null}) \!\leftarrow\! \mathrm{TextEnc}(\mathcal P)$, \ 
       $\mathbf e \!\leftarrow\! \mathrm{SampleCond}(\mathbf e_{pos},\mathbf e_{null})$
\State \textbf{UNet denoising:} 
       $\epsilon_{rgb}\!\leftarrow\!\mathcal{U}(z_{rgb}, t, \mathbf e)$, \ 
       $\epsilon_{ir}\!\leftarrow\!\mathcal{U}(z_{ir}, t, \mathbf e)$
\State \textbf{Latent update:} $\tilde z_{rgb}\!\leftarrow\! z_{rgb}-\epsilon_{rgb}$, \ 
      $\tilde z_{ir}\!\leftarrow\! z_{ir}-\epsilon_{ir}$
\State \textbf{Decode:} $\hat I_{rgb}\!\leftarrow\!\mathcal{V}.\mathrm{dec}(\tilde z_{rgb})$, \ 
      $\hat I_{ir}\!\leftarrow\!\mathcal{V}.\mathrm{dec}(\tilde z_{ir})$
\State \textbf{SAR-guided refinement (Alg.\ref{alg:sar_guided}):}
      $\hat I_{rgb}\!\leftarrow\!\mathrm{Cross attention SAR Fusion}(\hat I_{rgb}, S_1)$,\ 
      $\hat I_{ir}\!\leftarrow\!\mathrm{Cross attention SAR Fusion}(\hat I_{ir}, S_1)$
\State \textbf{Loss:} $\mathcal L $; \ \textbf{update} $\theta$
\State \Return $\hat I_{rgb}, \hat I_{ir}$
\end{algorithmic}
\end{algorithm}

The pseudocode of Cross-Attention Knowledge Constraint Module and Cross-attention SAR Fusion Module are shown as Algorithm \ref{alg:know_guieded} and \ref{alg:sar_guided}.

\begin{algorithm}[H]
\caption{Cross-Attention Knowledge Constraint Module}
\label{alg:know_guieded}
\begin{algorithmic}[1]
\Require Image latent $\mathbf z_{\text{img}}$; DEM $D$, LandCover $L$, Month $m$; projections $\mathrm{Proj}$, $\mathrm{Proj}^{-1}$; learnable scale $\gamma$
\Ensure Updated latent $\hat{\mathbf z}_{\text{img}}$
\State $\mathbf f_{\text{DEM}} \leftarrow \text{Enc}_{\text{dem}}(D)$;\quad
       $\mathbf f_{\text{LC}} \leftarrow \text{Enc}_{\text{lc}}(L)$;\quad
       $\mathbf f_{\text{month}} \leftarrow \text{Enc}_{\text{mon}}(m)$
\State $\mathbf z_{\text{aux}} \leftarrow \mathbf f_{\text{DEM}} + \mathbf f_{\text{LC}} + \mathbf f_{\text{month}}$ \Comment{Eq.(5)}
\State $Q \leftarrow \mathrm{Proj}(\mathbf z_{\text{img}})$;\quad
       $K \leftarrow \mathrm{Proj}(\mathbf z_{\text{aux}})$;\quad
       $V \leftarrow \mathrm{Proj}(\mathbf z_{\text{aux}})$ \Comment{Eq.(6)}
\State $\mathrm{Attn} \leftarrow \mathrm{softmax}\!\big(\frac{QK^{\top}}{\sqrt{d}}\big)\,V$ \Comment{Eq.(7)}
\State $\hat{\mathbf z}_{\text{img}} \leftarrow \mathbf z_{\text{img}} + \gamma \cdot \mathrm{Proj}^{-1}(\mathrm{Attn})$ \Comment{Eq.(8)}
\State \Return $\hat{\mathbf z}_{\text{img}}$
\end{algorithmic}
\end{algorithm}

\begin{algorithm}[H]
\caption{Cross-attention SAR Fusion Module}
\label{alg:sar_guided}
\begin{algorithmic}[1]
\Require RGB/IR image $I_{v} \in \mathbb{R}^{3\times H\times W}$, 
SAR image $I_{sar} \in \mathbb{R}^{2\times H\times W}$; 
projection $\phi_v$, $\phi_{sar}$; 
parameters $\gamma$, $W_q, W_k, W_v$
\Ensure Fused feature $F_{\text{fused}}$
\State $F_v \leftarrow \phi_v(I_v)$; \quad $F_{sar} \leftarrow \phi_{sar}(I_{sar})$ \Comment{Feature projection}
\State $Q \leftarrow W_q F_v$; \quad $K \leftarrow W_k F_{sar}$; \quad $V \leftarrow W_v F_{sar}$
\State $\text{Attn}(Q,K,V) \leftarrow \text{Softmax}\!\left(\frac{QK^\top}{\sqrt{d}}\right)V$
\State $F_{\text{fused}} \leftarrow F_v + \gamma \cdot G(F_{sar}) \odot \text{Attn}(Q,K,V)$ \Comment{Residual gated fusion}
\State \Return $F_{\text{fused}}$
\end{algorithmic}
\end{algorithm}

\section{Loss Curves}
\label{sec:curves}
\begin{figure}[H]
    \centering
    \begin{subfigure}{0.48\linewidth}
        \includegraphics[width=\linewidth]{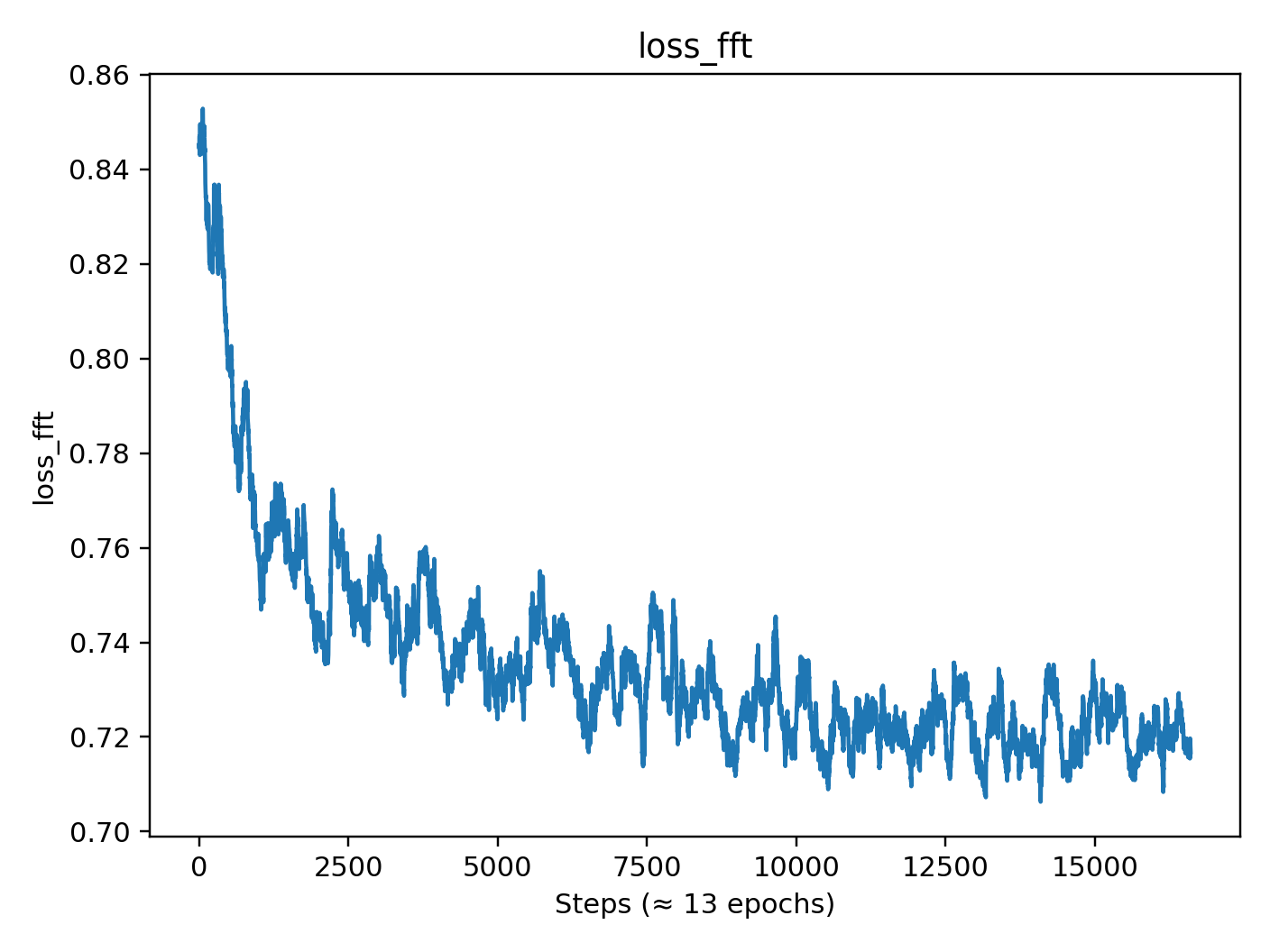}
        \caption{FFT loss}
    \end{subfigure}
    \begin{subfigure}{0.48\linewidth}
        \includegraphics[width=\linewidth]{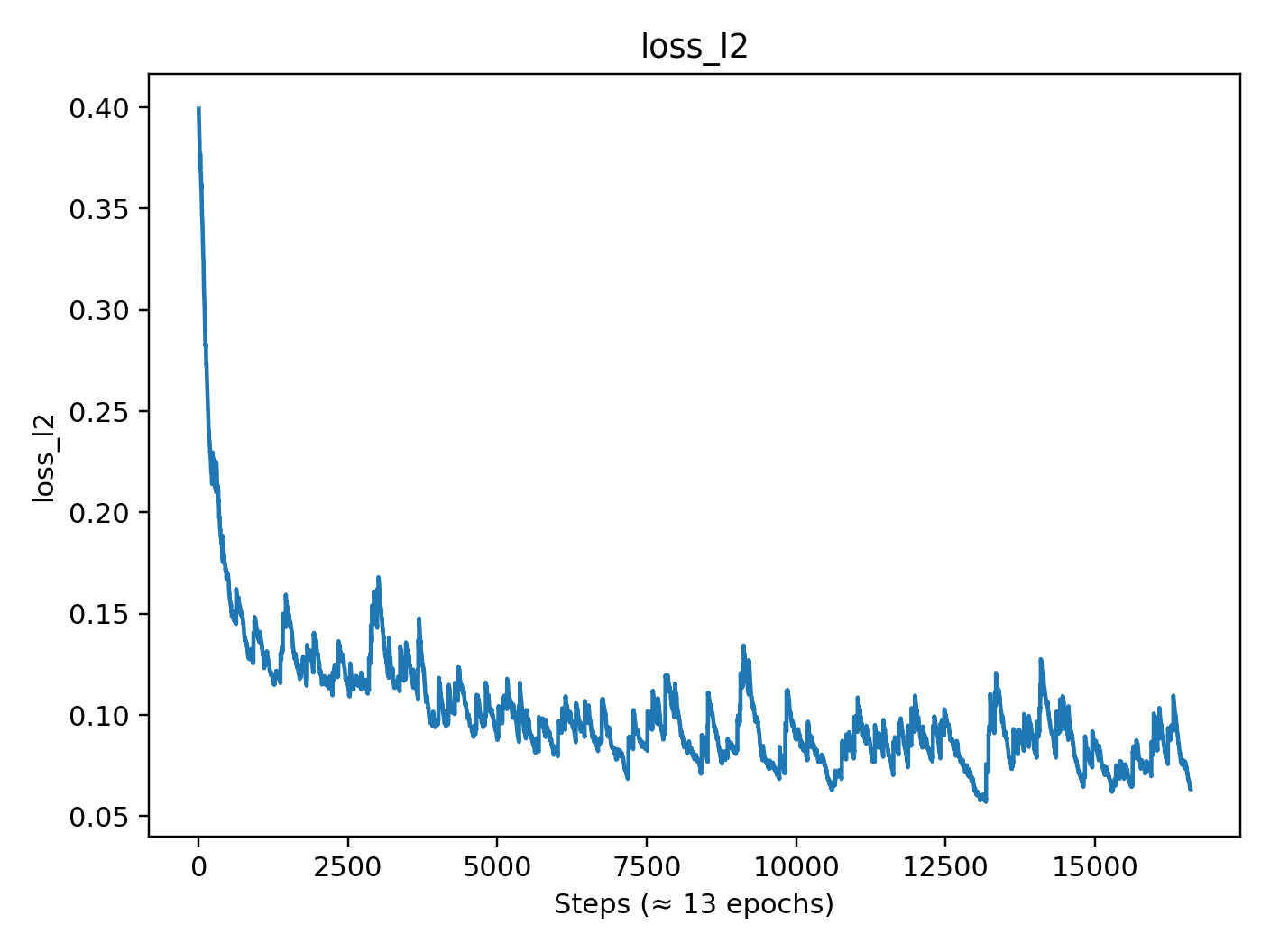}
        \caption{L2 loss}
    \end{subfigure}
    \begin{subfigure}{0.48\linewidth}
        \includegraphics[width=\linewidth]{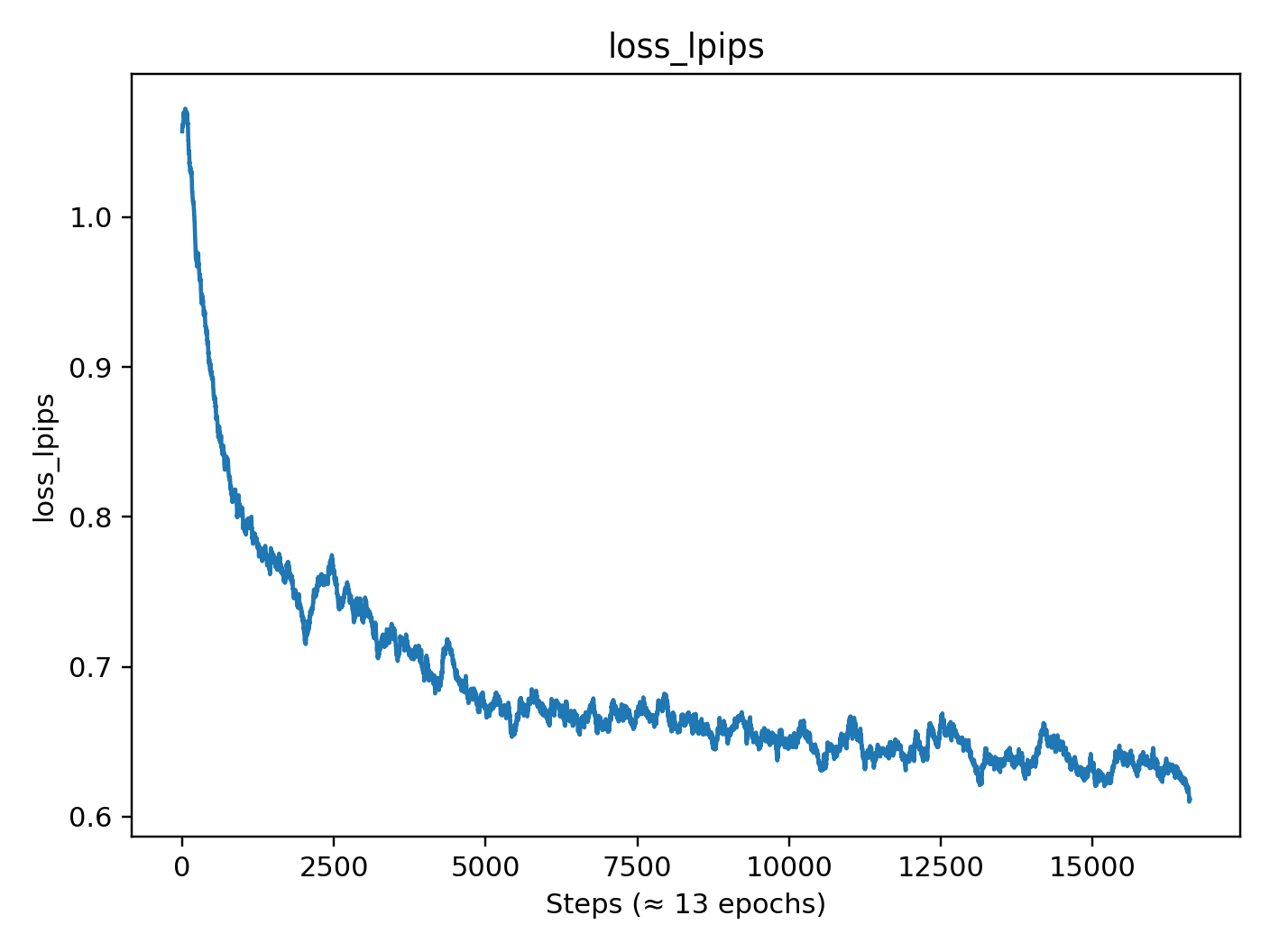}
        \caption{LPIPS loss}
    \end{subfigure}
    \begin{subfigure}{0.48\linewidth}
        \includegraphics[width=\linewidth]{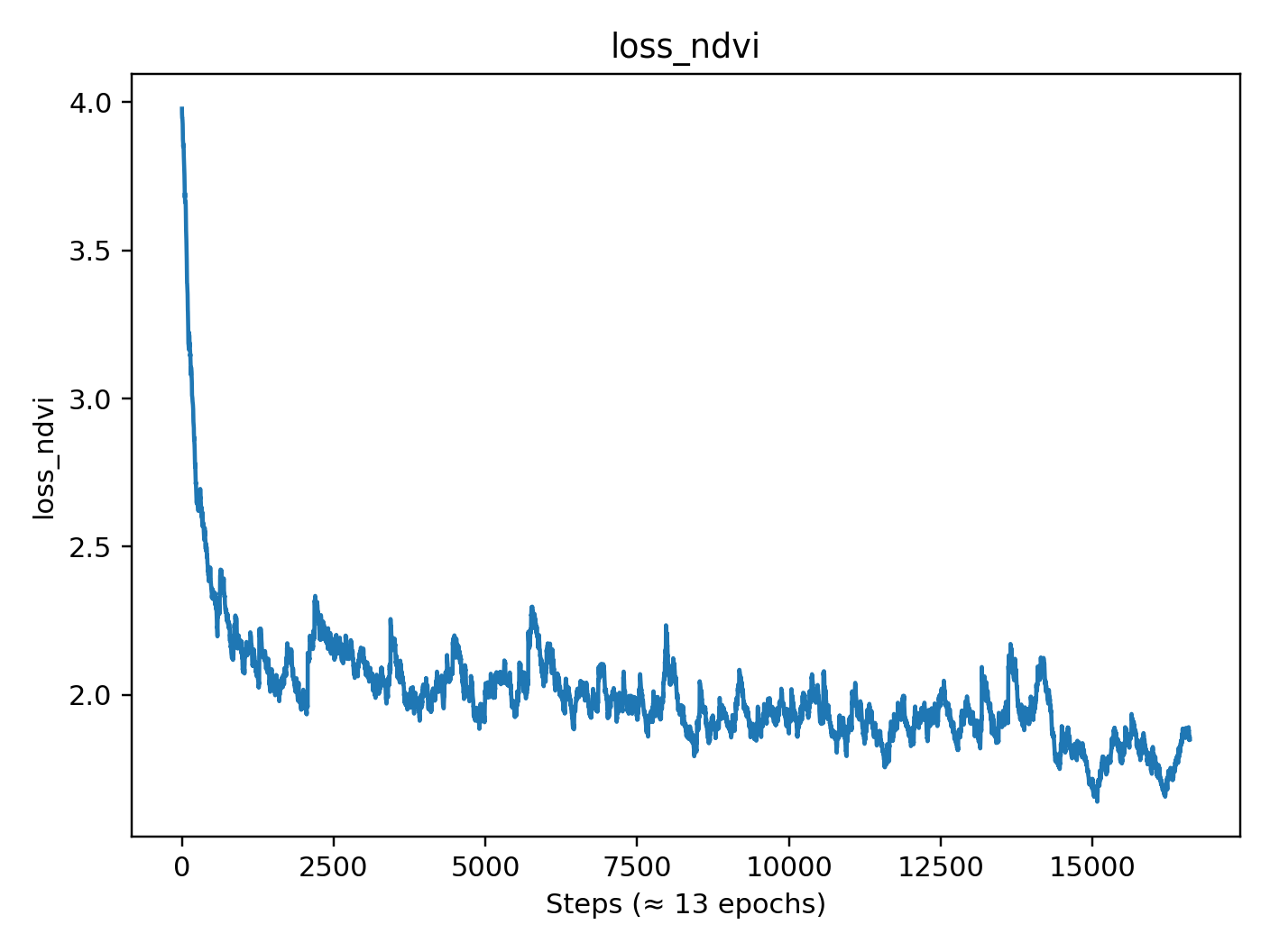}
        \caption{NDVI loss}
    \end{subfigure}
    \begin{subfigure}{0.48\linewidth}
        \includegraphics[width=\linewidth]{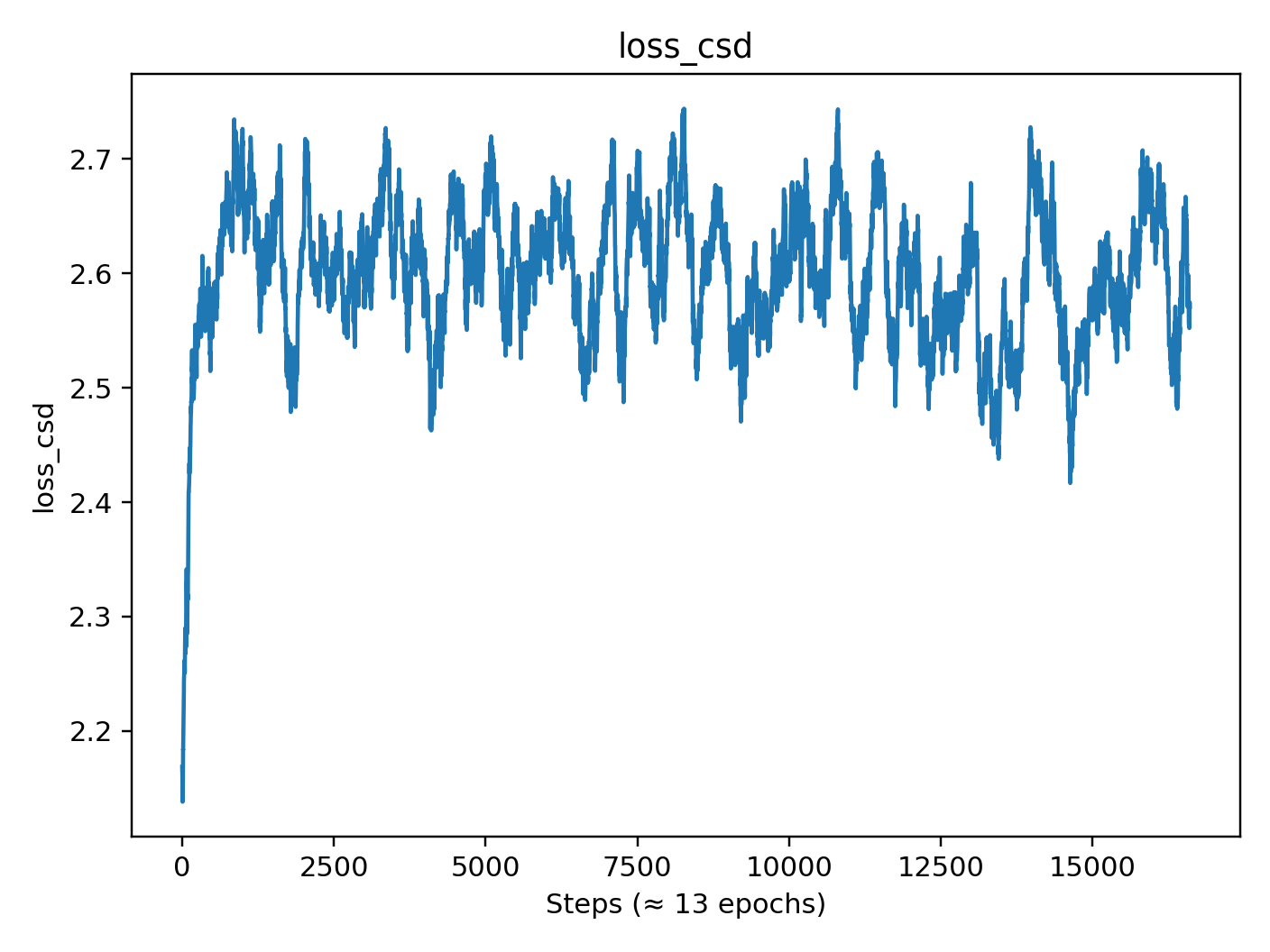}
        \caption{CSD loss}
    \end{subfigure}
    \caption{Training curves for individual loss components.}
    \label{fig:loss_curves}
\end{figure}

In this section, we provide training curves for individual loss components, in Figure \ref{fig:loss_curves}. The pixel-level losses (FFT, L2, NDVI) and perceptual loss (LPIPS) decrease steadily, indicating stable convergence of the reconstruction objective. Although the CSD loss was originally proposed in 3D generation tasks to optimize the posterior probability of rendered images aligning their semantic content with text prompts \cite{sun2025pisasr}, in our setting it is repurposed as a semantic consistency constraint across spatial-spectral domains rather than image–text alignment. Without the regularizing effect of text embeddings, the feature distributions exhibit larger variance across batches, which naturally leads to oscillations in the loss curve. However, the CSD loss remains statistically stable throughout training, indicating the convergence of the training process.

 \bibliographystyle{elsarticle-num} 
 \bibliography{cas-refs}





% \end{thebibliography}
\end{document}